\documentclass[letterpaper, 10 pt, conference]{ieeeconf}
\IEEEoverridecommandlockouts                              %

\usepackage{booktabs}
\usepackage{multirow}
\usepackage{graphicx}
\usepackage{amsmath}
\usepackage[bookmarks=true,colorlinks]{hyperref}

\newcommand{\ie}{\textit{i.e.}, }
\newcommand{\eg}{\textit{e.g.}, }
\newcommand{\mysubsection}[1]{\vspace{2mm}\noindent{\textbf{#1}}}
\DeclareMathOperator*{\argmax}{arg\,max}
\newcommand{\x}{\hphantom{1}}

\title{\LARGE \bf
Spatial Intention Maps for Multi-Agent Mobile Manipulation
}
\author{Jimmy Wu$^{1,2}$, Xingyuan Sun$^{1}$, Andy Zeng$^{2}$, Shuran Song$^{3}$, Szymon Rusinkiewicz$^{1}$, Thomas Funkhouser$^{1,2}$
\thanks{$^{1}$Princeton University, $^{2}$Google, $^{3}$Columbia University}%
}

\begin{document}

\maketitle
\thispagestyle{empty}
\pagestyle{empty}

\begin{abstract}

The ability to communicate intention enables decentralized multi-agent robots to collaborate while performing physical tasks. In this work, we present spatial intention maps, a new intention representation for multi-agent vision-based deep reinforcement learning that improves coordination between decentralized mobile manipulators. In this representation, each agent's intention is provided to other agents, and rendered into an overhead 2D map aligned with visual observations. This synergizes with the recently proposed spatial action maps framework, in which state and action representations are spatially aligned, providing inductive biases that encourage emergent cooperative behaviors requiring spatial coordination, such as passing objects to each other or avoiding collisions. Experiments across a variety of multi-agent environments, including heterogeneous robot teams with different abilities (lifting, pushing, or throwing), show that incorporating spatial intention maps improves performance for different mobile manipulation tasks while significantly enhancing cooperative behaviors.

\end{abstract}

\section{Introduction}

Multi-agent systems require a mutual understanding of intention in order to coordinate amongst each other and perform collaborative tasks.
This is particularly important for decentralized robots that share the same physical space while interacting in complex environments under partial observability and limited-bandwidth communication.
Applications of such systems include foraging~\cite{mataric1995issues}, hazardous waste cleanup~\cite{parker1998alliance}, object transportation~\cite{rus1995moving,tuci2018cooperative}, and search and rescue~\cite{murphy2000marsupial}.
The capacity to coordinate by conveying intentions can reduce collisions between robots (which can be both costly and disabling) while improving the efficiency with which tasks are completed.

Intention-aware multi-agent learning methods have shown promising success---from hide and seek~\cite{baker2019emergent} to Starcraft~\cite{vinyals2019grandmaster}---albeit largely in simulated domains such as video games.
These systems often communicate intentions via sharing high-level state information, or by condensing intentions into a low-dimensional embedding such as the coordinates of a destination.
Both of these approaches, however, exclude spatial structure and are thus a poor fit for reinforcement learning (RL) with convolutional neural networks, which has become a dominant strategy for learning with visual input.

\begin{figure}
\includegraphics[width=\columnwidth]{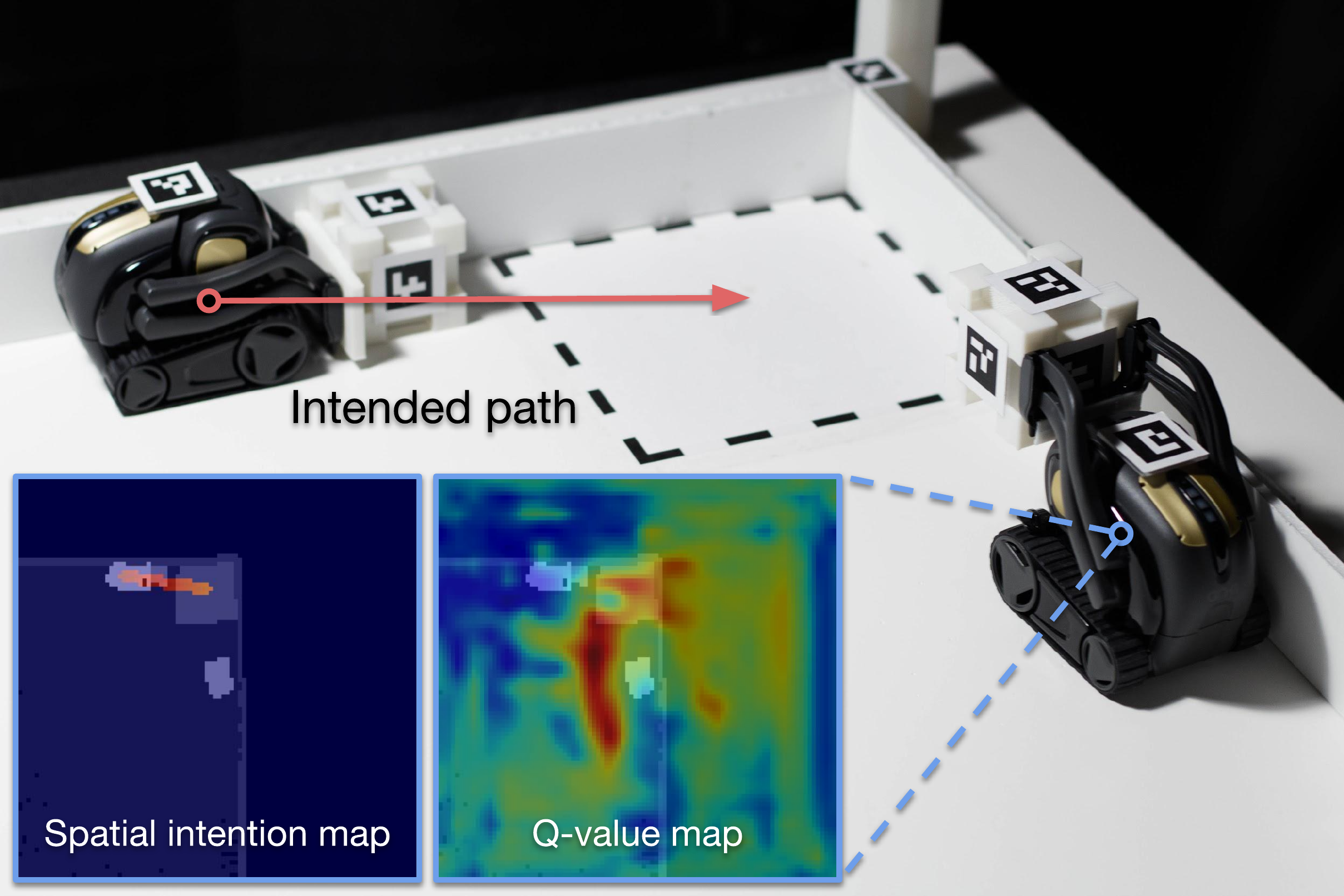}
\vspace{-6mm}
\caption{Spatial intention maps allow agents to choose actions with knowledge of the actions being performed by other agents. In the figure, the robot on the left is moving towards the upper right corner.
The robot on the right sees this in the spatial intention map, and instead of moving towards the goal straight ahead, it chooses to move to the left (dark red in the Q-value map).  This avoids a potential collision with the other robot.}
\vspace{-4mm}
\label{fig:teaser}
\end{figure}

In this work we propose \emph{spatial intention maps} (Fig.~\ref{fig:teaser}), a new intention representation for multi-agent deep RL.
In our approach, each agent's \emph{intention} (\ie its most recently selected action) is rendered into a 2D map  (first inset in Fig.~\ref{fig:teaser}) aligned with a state map derived from visual observations.
The intention and state maps are input into a fully convolutional network to predict Q-values for a dense set of spatial actions (second inset in Fig.~\ref{fig:teaser}).
At each step, the best action (\eg navigate to a location) is selected for execution.
Since spatial intention maps are encoded as images pixelwise aligned with state and action maps~\cite{wu2020spatial}, we expect that fully convolutional networks can leverage them more efficiently than other intention encodings.
Indeed, we find during experiments that encoding intentions spatially in this way provides a signal that encourages Q-functions to learn emergent collaborative behaviors, ranging from collision avoidance to spatial coordination and task specialization.

We demonstrate the efficacy of spatial intention maps across a variety of multi-agent environments, including robots with different abilities (lifting, pushing, or throwing) that are trained on different tasks (\emph{foraging} and \emph{search and rescue}).
We observe that using spatial intention maps leads to better performance and significant gains in collaborative behavior, including fewer collisions, coordination around bottlenecks such as doorways, and improved distribution of agents throughout an environment.
We also show that our learned policies (trained in simulation) are able to generalize to a real world setting.
For qualitative video results and code, please see the supplementary material at \url{https://spatial-intention-maps.cs.princeton.edu}.

\begin{figure*}
\includegraphics[width=\textwidth]{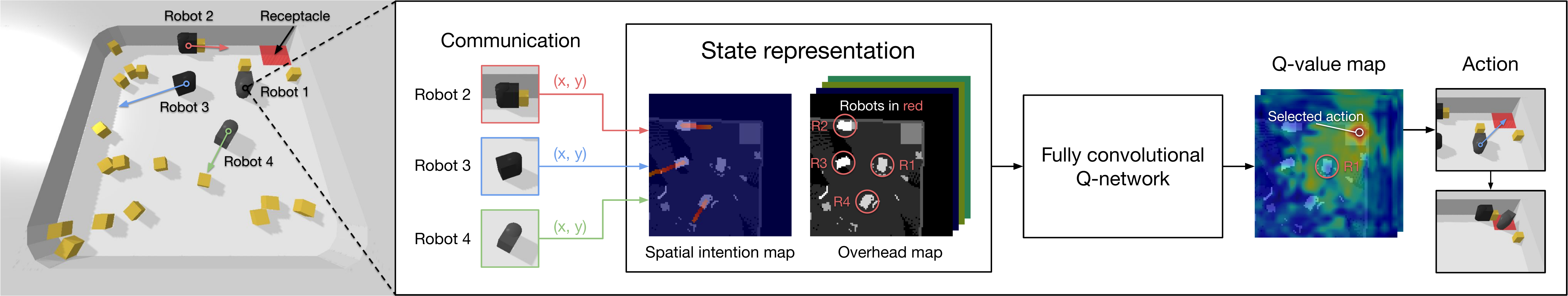}
\vspace{-6mm}
\caption{\textbf{Overview.}
Our system uses decentralized, asynchronous execution with communicating agents.
In the figure, robot 1 is currently choosing a new action.
It receives the intentions (represented as path waypoints) of the other agents, which are then encoded into a spatial map and used as input to the fully convolutional DQN policy network.}
\vspace{-4mm}
\label{fig:overview}
\end{figure*}

\section{Related Work}

\vspace{-2mm}\mysubsection{Multi-robot systems.} The earliest works in multi-robot systems date back to the 1980s~\cite{parker2016multiple}.
These works have studied a wide variety of challenges in multi-robot systems, such as architecture, communication, team heterogeneity, and learning.
Typical applications areas include foraging~\cite{mataric1995issues}, hazardous waste cleanup~\cite{parker1998alliance}, object transportation~\cite{rus1995moving,tuci2018cooperative}, search and rescue~\cite{murphy2000marsupial}, and soccer~\cite{kitano1997robocup}.
Many of these early systems used reactive or behavior-based approaches, which require hand-crafting policies for the agents.
In contrast to these works, we use reinforcement learning to enable our robots to automatically learn behaviors through experience.

\mysubsection{Multi-agent reinforcement learning.} There is a wide body of work in multi-agent reinforcement learning (MARL)~\cite{weiss1999multiagent,stone2000multiagent,busoniu2008comprehensive}.
One of the earliest works in this area was \cite{tan1993multi}, in which multiple independent Q-learning agents were trained in a cooperative setting.
More recently, with the rising popularity of deep reinforcement learning approaches such as DQN~\cite{mnih2015human}, many works have proposed ways to extend these methods to multi-agent settings~\cite{hernandez2019survey}.
These include addressing nonstationarity through modifications to experience replay~\cite{foerster2017stabilising,omidshafiei2017deep} or training with a centralized critic~\cite{lowe2017multi,foerster2017counterfactual}, or improving credit assignment through decomposing the value function~\cite{sunehag2017value,rashid2018qmix,zhang2020multi}.
Most of these approaches assume access to the full state of the environment, though a few use raw visual data~\cite{tampuu2017multiagent,jaderberg2019human}. In our approach, we learn directly from visual data reconstructed from partial observations of the environment.

\mysubsection{Learning-based multi-robot systems.} While there is a vast amount of literature on multi-agent reinforcement learning, most works focus on games and simulation environments. There is comparatively less work applying multi-agent learning to robotic systems.
Some earlier works have applied Q-learning to applications such as box pushing~\cite{mahadevan1992automatic}, foraging~\cite{mataric1997reinforcement}, soccer~\cite{asada1994coordination,stone1998towards}, and multi-target observation~\cite{fernandez2005reinforcement}.
More recently, \cite{xiao2020macro} and \cite{xiao2020learning} have used macro-actions~\cite{amato2019modeling} with DQN to enable multiple robots to asynchronously reason and execute actions.
Other works have investigated multi-agent tasks such as navigation~\cite{wang2020cooperation} or cooperative manipulation~\cite{nachum2020multi}.
As with the MARL works previously described, these typically assume access to high-level state information such as the positions of relevant objects.
In contrast, we learn directly from visual data, allowing agents to automatically learn to detect relevant visual features.

\mysubsection{Multi-robot communication.} Many works in multi-robot systems have studied the effect of communication on improving team performance~\cite{maclennan1993synthetic,balch1994communication,parker1995effect}.
These comparisons typically span a continuum from no communication (implicit), to passive observation of teammates' states, to direct communication (explicit).
The same continuum of communication has been studied for multi-agent learning~\cite{tan1993multi,zhang2013coordinating,panait2005cooperative}.
More recent works in multi-agent learning have studied other ways to communicate, such as learning what to  communicate~\cite{foerster2016learning,sukhbaatar2016learning,mordatch2017emergence} or learning to model the intentions of other agents~\cite{stulp2006implicit,qi2018intent,jaques2019social,ndousse2020multi,wang2020cooperation}.
We similarly explore this communication continuum, but in our approach, communicated intentions are spatially encoded and aligned with the state and action representations.
This allows agents to reason about their teammates' intentions in the same domain as their own observations and actions.

\section{Method}

We study spatial intention maps in the context of two tasks.
The first is a \emph{foraging} task, in which a team of robots work together to move all objects in an environment to a target receptacle in the upper right corner (see Fig.~\ref{fig:overview} for an example environment).
The second is a \emph{search and rescue} task, in which a team of robots need to find and ``rescue'' all objects in an environment.
We train our policies in a PyBullet~\cite{coumans2016pybullet} simulation environment and run them on real robots using a sim-to-real mirroring setup (see Sec.~\ref{sec:experiments-real} for more details).

\subsection{Reinforcement Learning Formulation}

We model our task as a Markov decision process (MDP) from the perspective of each individual agent, using single-agent state and action spaces. We train our policies using double deep Q-learning (DQN)~\cite{mnih2015human,van2016deep}, giving individual rewards to agents during training.
This formulation of multi-agent RL is similar to independent Q-learning~\cite{tan1993multi}, but we share policies between agents of the same type during training.
Execution is decentralized---trained policies are run independently and asynchronously on each agent.

As widely discussed in the multi-agent literature, a common challenge with training multiple agents concurrently in the same environment is nonstationarity caused by other agents having policies that are constantly evolving. In our system, we use spatial intention maps (see Fig.~\ref{fig:overview}) to directly encode the intentions of other agents into the state representation given to DQN, which helps to alleviate the uncertainty in other agents' behaviors. Please see the supplementary material for more training details.

\subsection{State Representation}
\label{sec:methods-state-representation}

Our state representation consists of a local overhead map, 
along with a series of auxiliary local maps with additional information useful to the agent when making its decisions~\cite{wu2020spatial}.
To construct these local maps, each agent independently builds up its own global map of the environment using online mapping.
Each agent has a simulated forward-facing RGB-D camera and captures partial observations of the environment, which are integrated with previous views to gradually build up a global overhead map.
Agents must therefore learn policies that can deal with outdated information and seek out unexplored parts of the environment.

Each time an agent chooses a new action, it crops out a local map from its global map to generate a new state representation.
These local maps are oriented such that the agent itself is in the center and facing up (see Fig.~\ref{fig:overview} for an example).
The state representation consists of the following local maps, each of which is represented as an overhead image:
(i) an environment map,
(ii) an agent map encoding the agent's own state and the observed states of other agents,
(iii-iv) maps encoding shortest path distances to the receptacle and from the agent, and
(v) the spatial intention map, which we describe in Sec.~\ref{sec:methods-spatial-intention-maps}.
The agent map spatially encodes the state of every agent, which includes each agent's pose, as well as whether they are carrying an object (if applicable).

\subsection{Action Representation}

We use spatial action maps~\cite{wu2020spatial} as our action representation.
The action space is represented as a pixel map, spatially aligned with the state representation, in which every pixel represents the action that navigates the agent to the corresponding location in the environment (see Q-value map in Fig.~\ref{fig:overview}).
For agents that can execute an end effector action (such as lifting or throwing), we augment the action space with a second spatial channel, which represents navigating to the corresponding location and then attempting to perform the end effector action.
(Note that figures in this paper only show one relevant action channel for clarity.)
We execute the action corresponding to the argmax across all channels in the action space, employing high-level motion primitives implemented with low-level control.
For movement, the primitive attempts to proceed to the specified location along the shortest path, computed using the agent's own occupancy map.
For end effector actions, the primitives attempt to lock onto an object before operating the end effector.

\subsection{Spatial Intention Maps}
\label{sec:methods-spatial-intention-maps}

Spatial intention maps provide an encoding of other agents' intentions in the form of a map aligned with the state and action representations. This map-based encoding of intentions is key to our approach, as it allow us to train a fully convolutional deep residual Q-network~\cite{long2015fully,he2016deep} that maps pixels representing states (and intentions) to a Q-value map that is pixelwise aligned with the state. This type of computation (dense pixelwise prediction using fully convolutional networks) has been shown to be effective for many visual tasks (\eg semantic segmentation).  In our case, it is a good fit because encoding intentions into the spatial domain allows deep Q-networks to reason about intentions in the same domain as the state representation.

Agents in our setup use decentralized, asynchronous execution, which means that whenever an agent is choosing a new action, all of the other agents in the environment will be in motion, executing their most recently selected action.  We encode these in-progress actions spatially as rasterized paths in a spatial intention map.  Intended paths are encoded using a linear ramp function, with a value of 1 at the executing agent's current location, and dropping off linearly along the path (see Fig.~\ref{fig:overview} for an example).
A lower value at a point on a path indicates a longer period of time before the executing agent will reach that point.
This information enables more fine-grained reasoning about time and distance (\eg the agent over there intends to come here, but is still quite a distance away).

The bandwidth requirements for spatial intention maps during decentralized execution are low, as agents do not communicate maps (images) directly.
Instead, they broadcast intentions as lists of $(x, y)$ coordinates (waypoints of intended path).
Then, whenever an agent is selecting a new action, it locally renders the most recently received intentions (paths) into an up-to-date spatial intention map.

\section{Experiments}

\begin{figure}
\begin{center}
\setlength\tabcolsep{1pt}
\begin{tabular}{ccc}
\includegraphics[width=0.325\columnwidth]{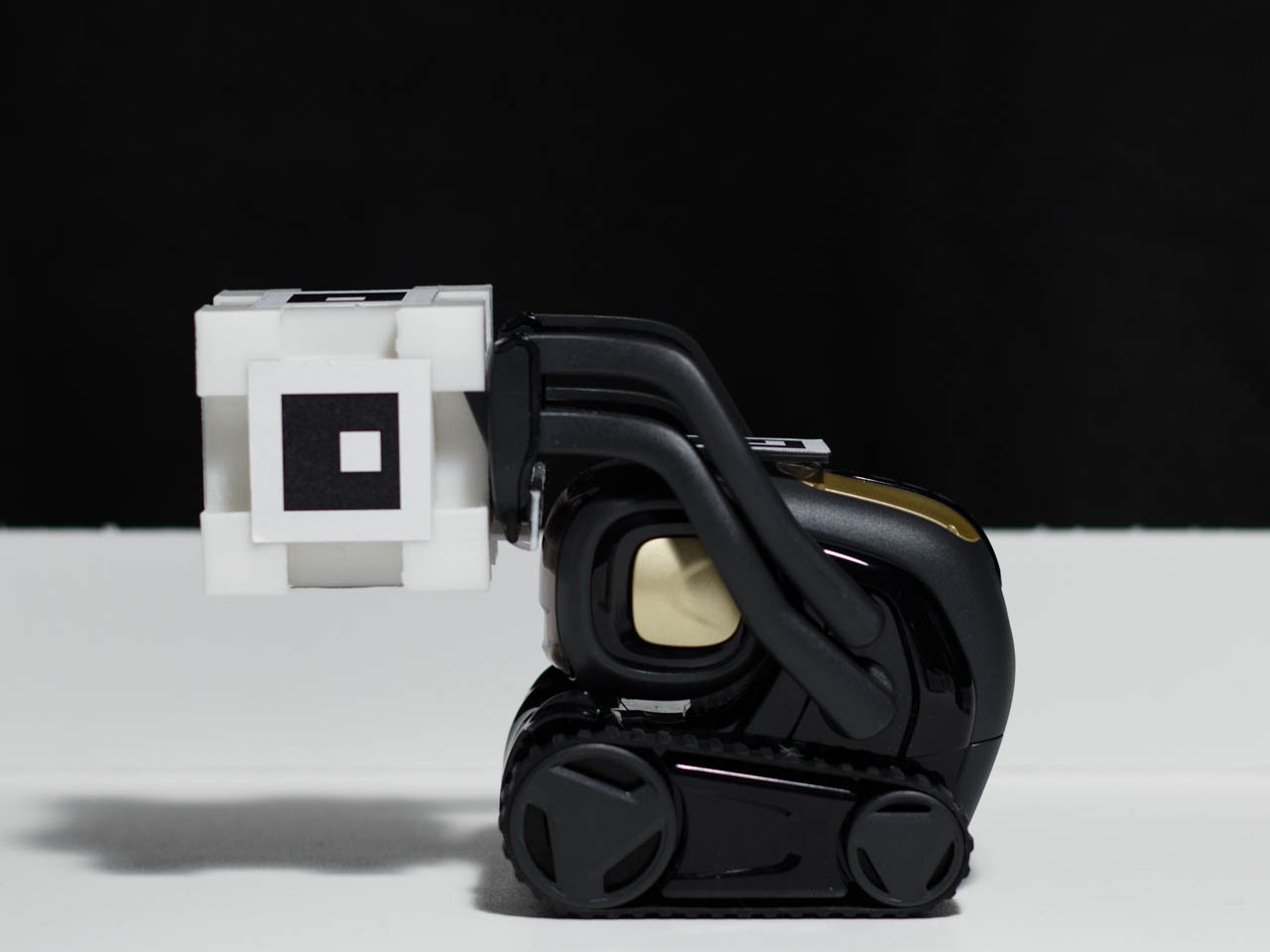} &
\includegraphics[width=0.325\columnwidth]{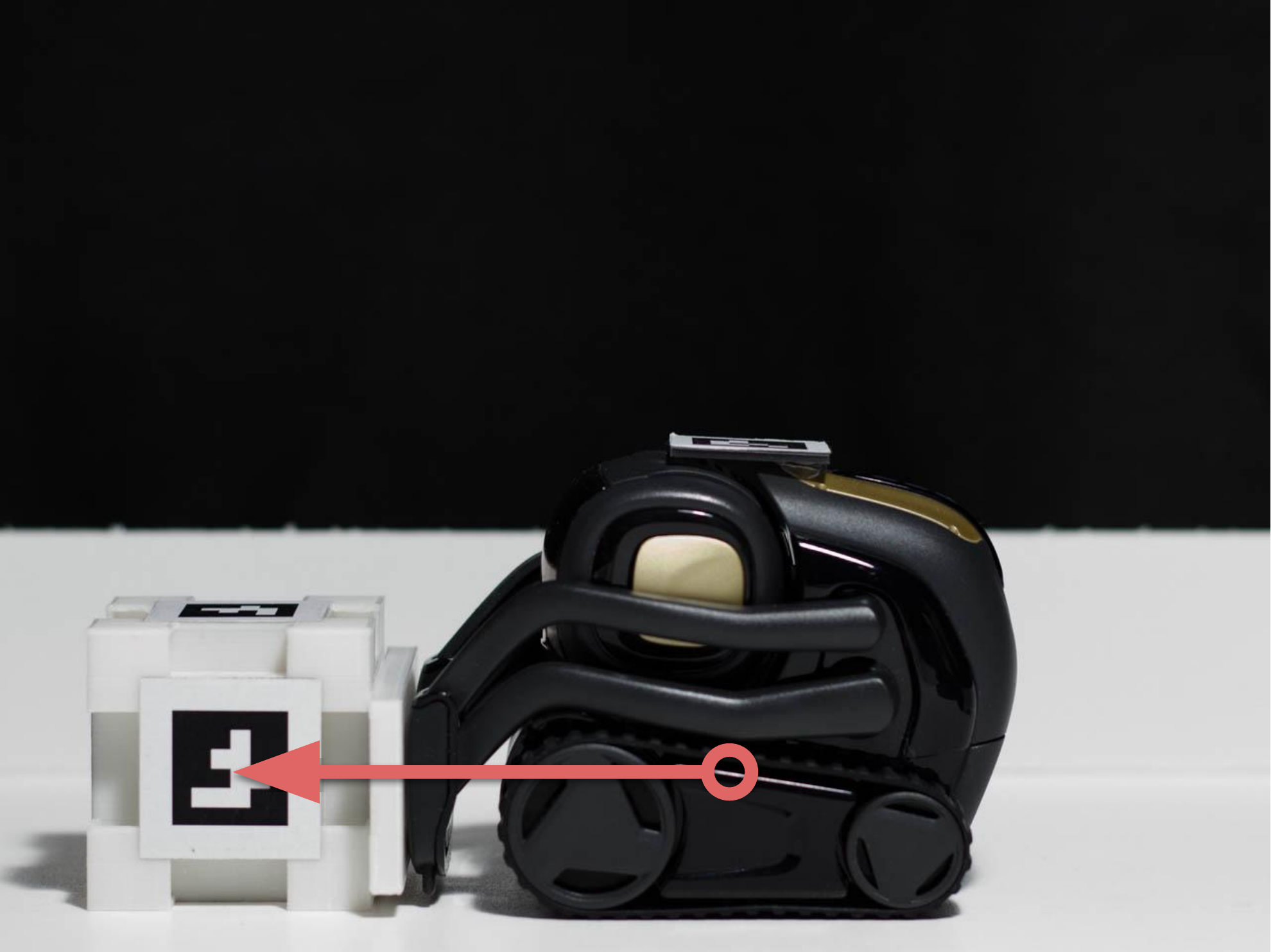} &
\includegraphics[width=0.325\columnwidth]{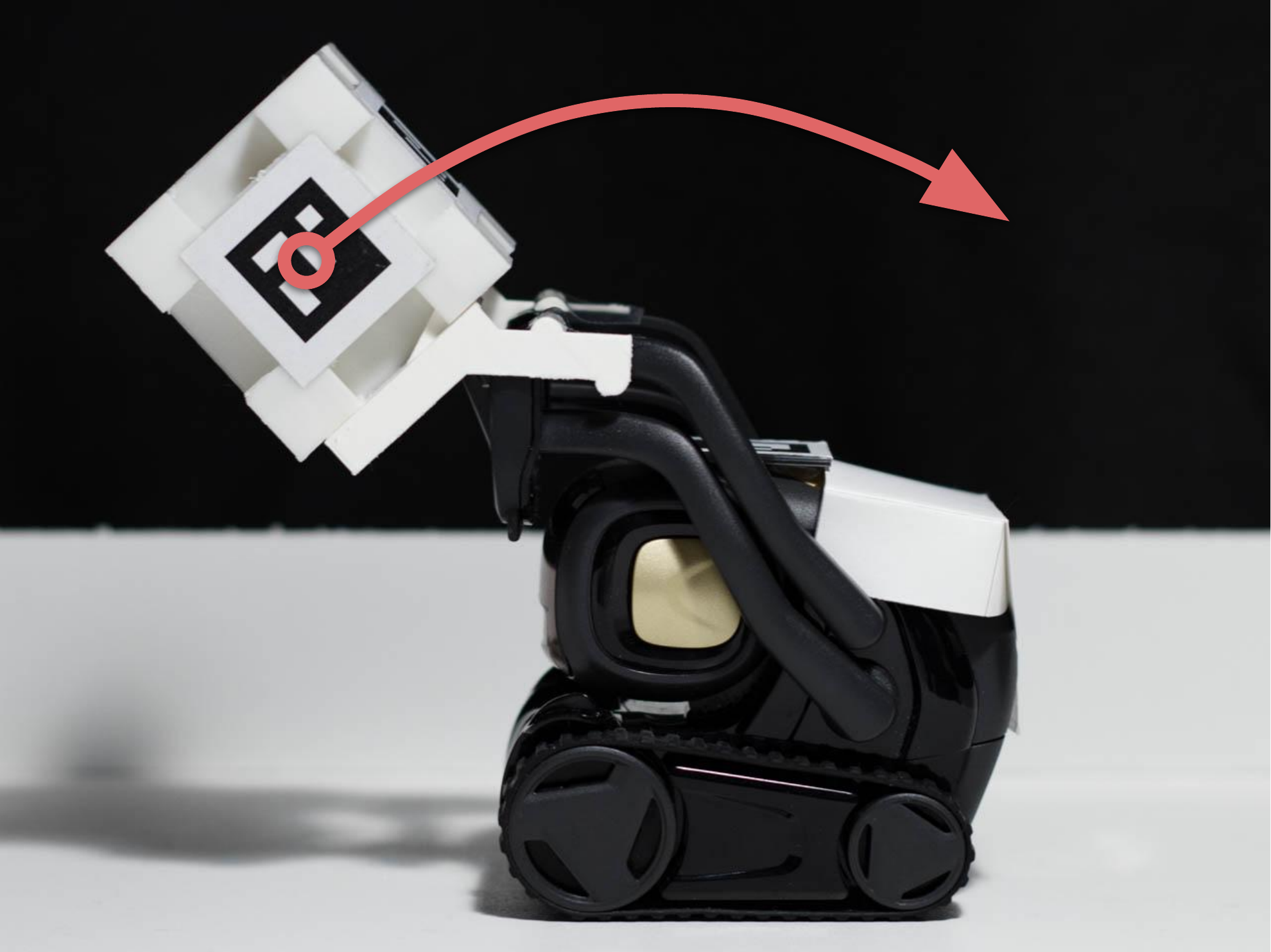} \\
\small{Lifting} & \small{Pushing} & \small{Throwing} \\
\end{tabular}
\end{center}
\vspace{-3mm}
\caption{We experiment with multiple robot types, each with a unique ability.}
\vspace{-4mm}
\label{fig:robot-types}
\end{figure}

We evaluate spatial intention maps with decentralized multi-robot teams on both \emph{foraging} and \emph{search and rescue} tasks across a wide range of simulation environments.
The goals of these experiments are to investigate
(i) whether learning with spatial intention maps is better than without, and
(ii) how learning with spatial intention maps compares with alternative representations of intention.
We also test our policies, which were trained in simulation, on real robots.

\subsection{Experimental Setup}

We investigate two tasks: (i) \emph{foraging}: where a team of robots must collaborate to move a set of objects into a receptacle (objects are removed from the environment when they enter the receptacle), and (ii) \emph{search and rescue}: where a team of robots must navigate to objects scattered in an environment and ``rescue'' them (objects are removed after contact by a robot).
The two tasks are trained in the same way for the most part.
Robots receive a success reward (+1.0) whenever an object is removed from the environment, and receive penalties for collisions with obstacles (-0.25) or other agents (-1.0).
In the foraging task, they receive a distance-based partial reward/penalty for moving an object closer to or further from the receptacle. The lifting robot is also penalized (-0.25) for dropping objects outside the receptacle.

We experiment with four types of robots (Fig. \ref{fig:robot-types}): (i) a lifting robot that can pick up objects and carry them, (ii) a pushing robot that can push objects, (iii) a throwing robot that can throw objects backwards, and (iv) a rescue robot that can mark an object as ``rescued'' upon contact.
Teams can be homogeneous (4 of the same) or heterogeneous combinations (2 robot types with 2 per type).
For all robots, an action consists of moving to a target location and optionally executing an end effector action (lifting or throwing).

\begin{figure}
\begin{center}
\setlength\tabcolsep{1pt}
\begin{tabular}{ccc}
\includegraphics[width=0.325\columnwidth]{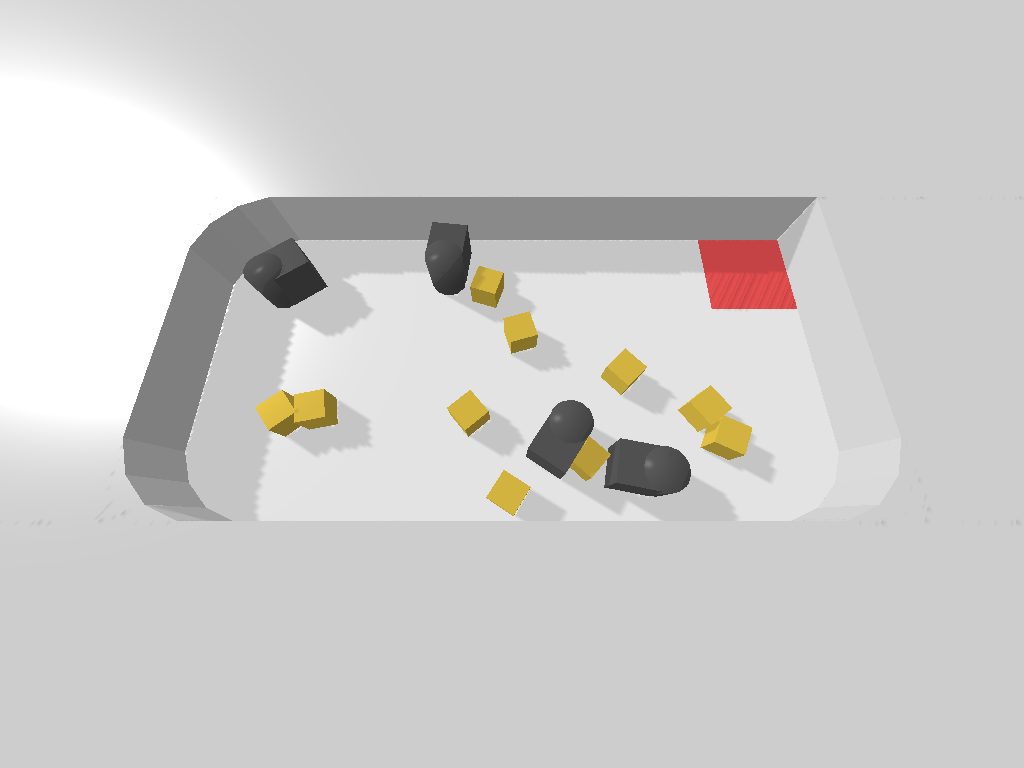} &
\includegraphics[width=0.325\columnwidth]{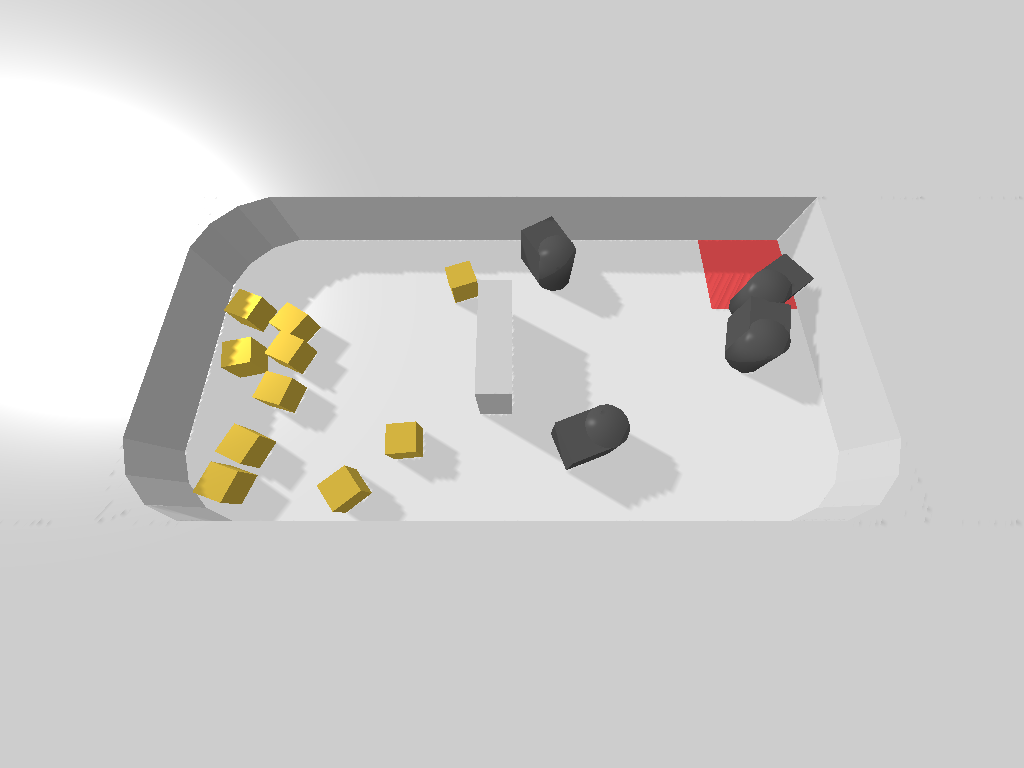} &
\includegraphics[width=0.325\columnwidth]{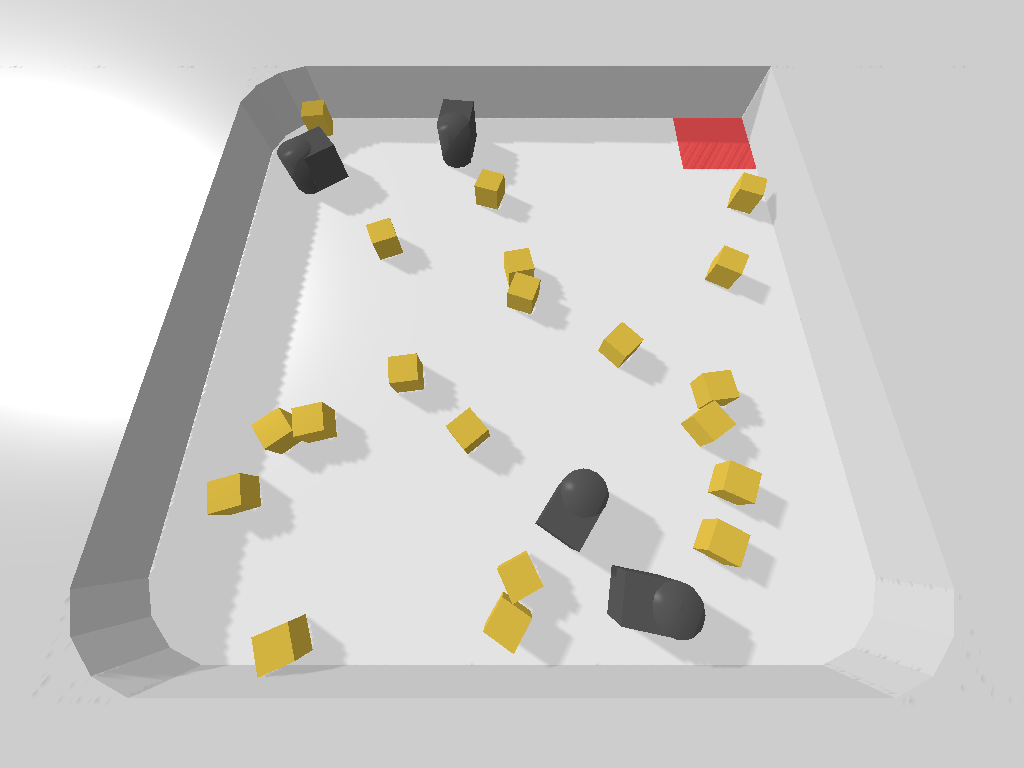} \\
\small{SmallEmpty} & \small{SmallDivider} & \small{LargeEmpty} \\
\vspace{-2mm} \\
\includegraphics[width=0.325\columnwidth]{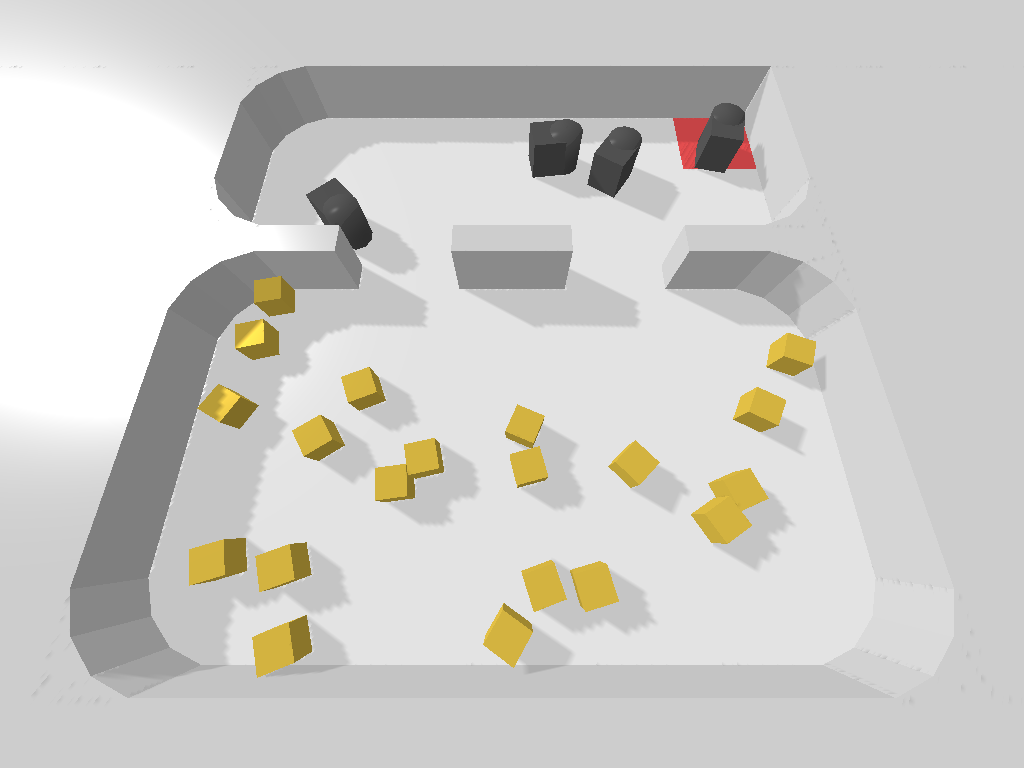} &
\includegraphics[width=0.325\columnwidth]{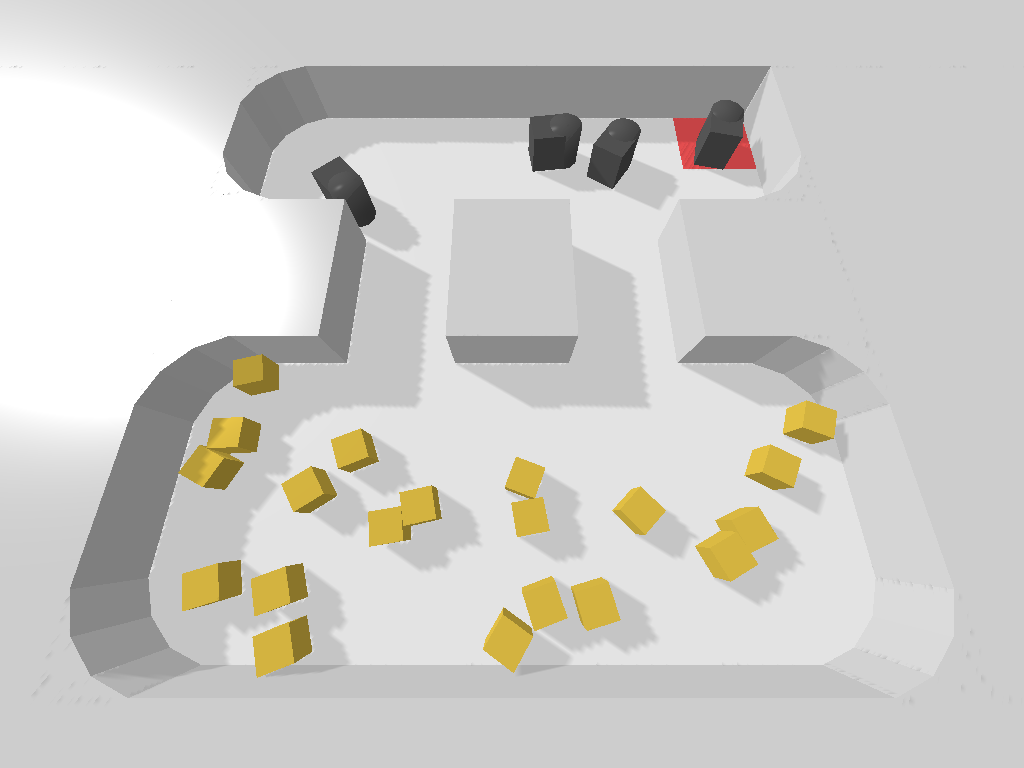} &
\includegraphics[width=0.325\columnwidth]{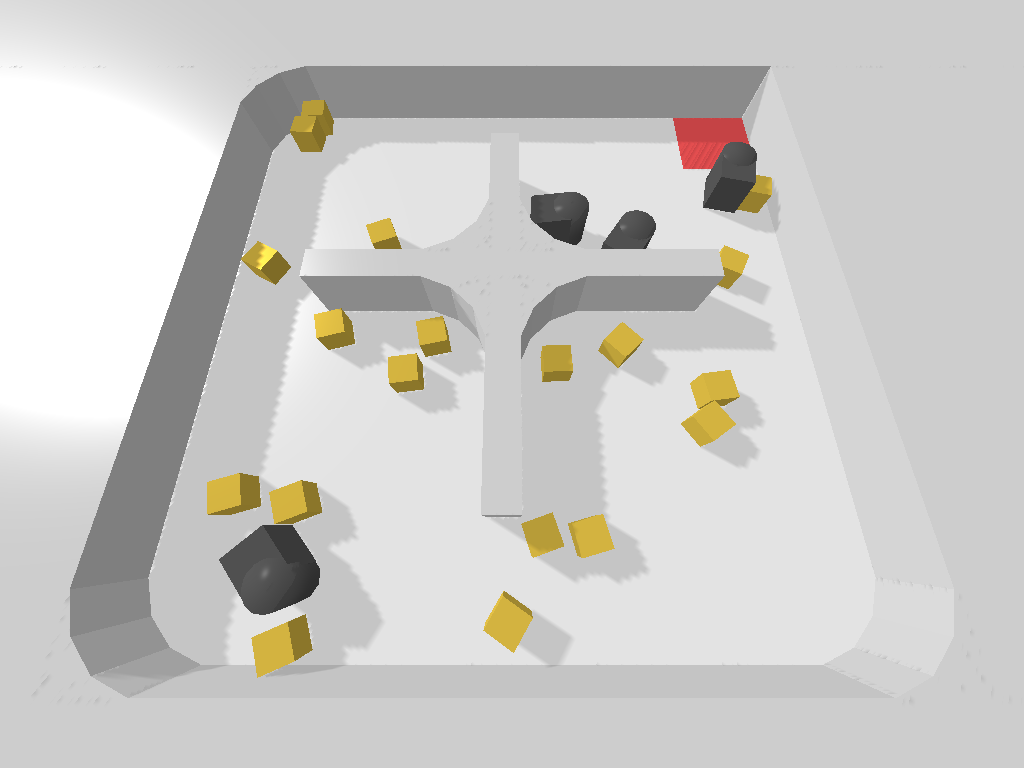} \\
\small{LargeDoors} & \small{LargeTunnels} & \small{LargeRooms} \\
\end{tabular}
\end{center}
\vspace{-2mm}
\caption{\textbf{Environments.} We ran experiments in six different environments with a variety of different obstacle configurations. In each environment, a team of robots (dark gray) is tasked with moving all objects (yellow) to the receptacle in the corner (red).}
\label{fig:envs}
\end{figure}

We test the robot teams in six different environments (see Fig.~\ref{fig:envs}).
In empty environments with no obstacles, robots need to coordinate to avoid colliding while going towards the receptacle.
In environments with obstacles, there are doorways and tunnels wide enough for only one robot to pass through, requiring coordination to avoid obstructing one another.
We initialize robots, objects, and obstacles (dividers and walls) in random configurations at the start of each episode.
Small environments contain 10 objects, and large ones contain 20.
For the SmallDivider, LargeDoors, and LargeTunnels environments, robots and objects are initialized in opposite sides of the room, requiring the robots to bring all objects through doorways or tunnels to reach the receptacle.

For each episode, our evaluation metric is the total number of objects gathered after a fixed time cutoff, which measures the efficiency of a team (higher is better).
The cutoff is the time at which the most efficient policy places the last object into the receptacle, and is different for every unique combination of robot team and environment, but is kept consistent between methods.
We evaluate a trained policy by averaging its performance over 20 test episodes.
For each method, we train five policies and report the mean and standard deviation of those averages across all five.
Between evaluation runs, randomly initialized environments are kept consistent using random seeds (training is not seeded).

\subsection{Experimental Results}
\label{sec:experiments-spatial-intention-maps}

In this section, we investigate whether spatial intention maps are useful by comparing the performance of multi-robot teams trained with and without them.

\mysubsection{Foraging task.}
We first investigate performance on the foraging task with teams of 4 lifting robots (4L) or 4 pushing robots (4P).
The corresponding rows in Tab.~\ref{tab:foraging} show a comparison of teams trained with and without spatial intention maps.
The results suggest that teams trained with spatial intention maps (Ours column) are able to learn the foraging task better.
The biggest quantitative differences occur in the larger and more complex environments---\eg the teams of lifting robots pick up an average of over 19 objects in the large environments with spatial intention maps, and only 11 to 17 without.
However, the differences are significant in all settings.

\begin{table}
\caption{Foraging Performance}
\vspace{-4mm}
\begin{center}
\begin{tabular}{l|l|cc}
\toprule
Robots & Environment & Ours & No intention maps \\
\midrule
4L & SmallEmpty & \textbf{\x9.54 $\pm$ 0.16} & \x7.92 $\pm$ 0.86 \\
& SmallDivider & \textbf{\x9.44 $\pm$ 0.46} & \x8.07 $\pm$ 0.75 \\
& LargeEmpty & \textbf{18.86 $\pm$ 0.85} & 15.58 $\pm$ 3.80 \\
& LargeDoors & \textbf{19.57 $\pm$ 0.25} & 13.96 $\pm$ 2.32 \\
& LargeTunnels & \textbf{19.00 $\pm$ 0.47} & 11.89 $\pm$ 5.96 \\
& LargeRooms & \textbf{19.52 $\pm$ 0.38} & 16.56 $\pm$ 1.53 \\
\midrule
4P & SmallEmpty & \textbf{\x9.51 $\pm$ 0.20} & \x8.73 $\pm$ 0.54 \\
& SmallDivider & \textbf{\x9.50 $\pm$ 0.24} & \x8.40 $\pm$ 0.78 \\
& LargeEmpty & \textbf{19.51 $\pm$ 0.53} & 18.86 $\pm$ 0.72 \\
\midrule
2L+2P & LargeEmpty & \textbf{19.52 $\pm$ 0.17} & 16.51 $\pm$ 4.27 \\
& LargeDoors & \textbf{19.55 $\pm$ 0.18} & 17.44 $\pm$ 0.63 \\
& LargeRooms & \textbf{19.51 $\pm$ 0.24} & 18.51 $\pm$ 0.75 \\
\midrule
2L+2T & LargeEmpty & \textbf{19.51 $\pm$ 0.67} & 12.46 $\pm$ 4.34 \\
& LargeDoors & \textbf{19.50 $\pm$ 0.45} & \x6.21 $\pm$ 4.12 \\
\bottomrule
\end{tabular}
\end{center}
\begin{center}
L = lifting, P = pushing, T = throwing
\end{center}
\vspace{-6mm}
\label{tab:foraging}
\end{table}

We conjecture one reason for the differences is that robots without intention maps tend to move conservatively back and forth to minimize collisions with nearby robots.
This is particularly evident when multiple robots are trying to approach the same object, or trying to put objects into the receptacle at the same time.
For example, they may all approach the receptacle, observe that others are also moving forward, then all preemptively back away.
In contrast, with intention maps, each robot is able to consider the intentions of other robots in order to choose actions that do not conflict.

\begin{figure}
\begin{center}
\setlength\tabcolsep{0pt}
\begin{tabular}{cc@{\hspace{1pt}}c@{\hspace{1pt}}cc}
\includegraphics[width=0.248\columnwidth]{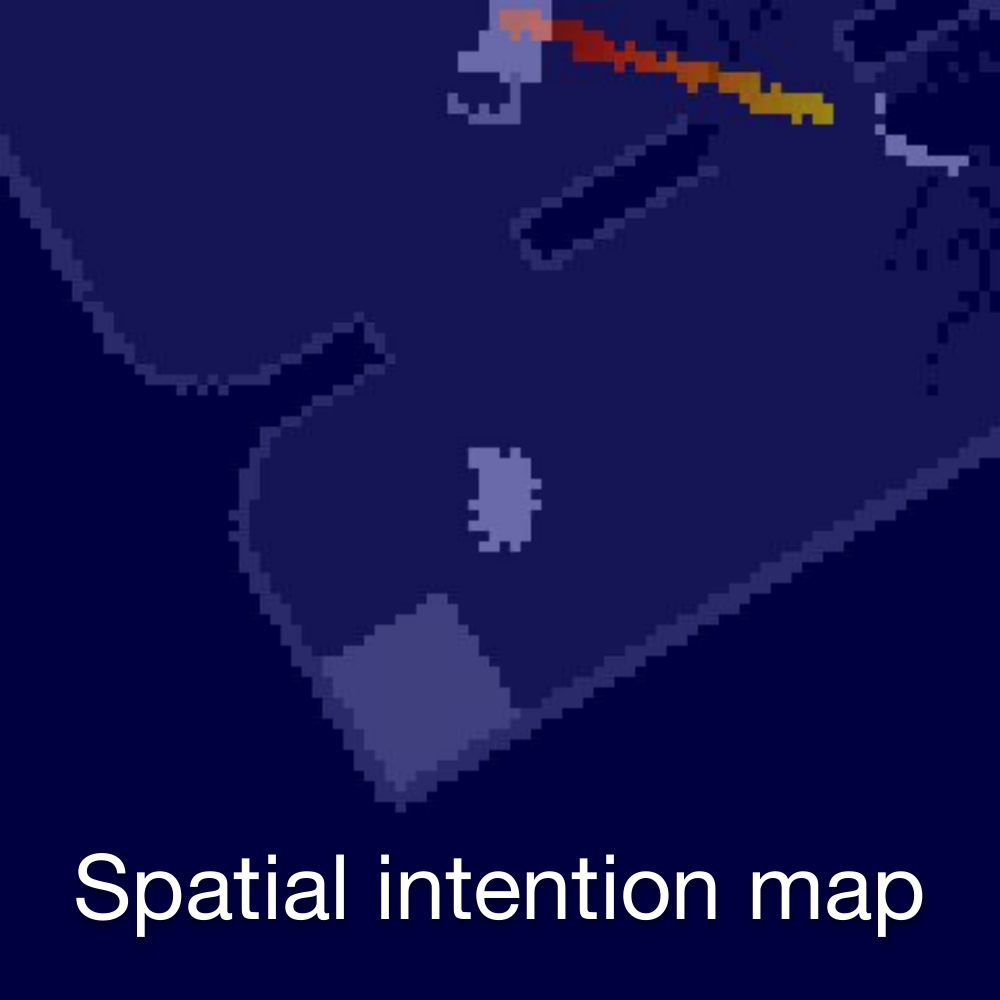} &
\includegraphics[width=0.248\columnwidth]{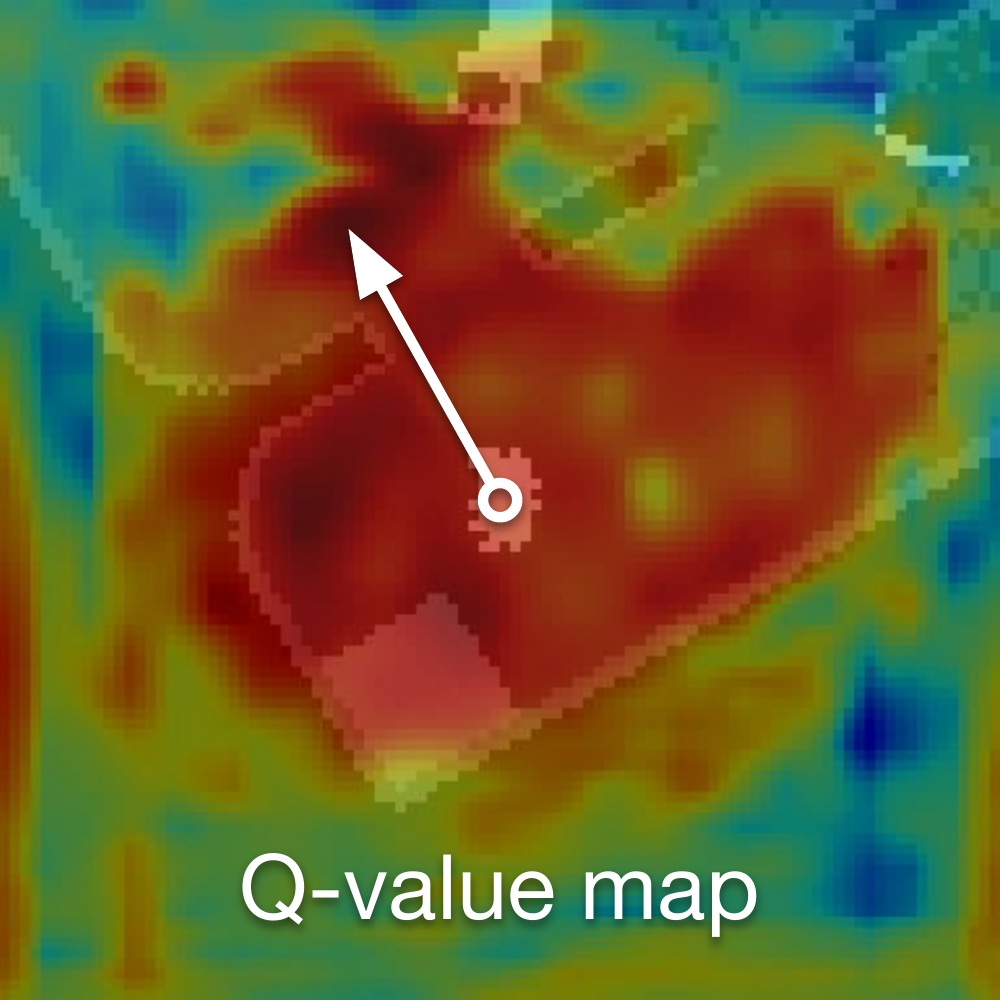} & &
\includegraphics[width=0.248\columnwidth]{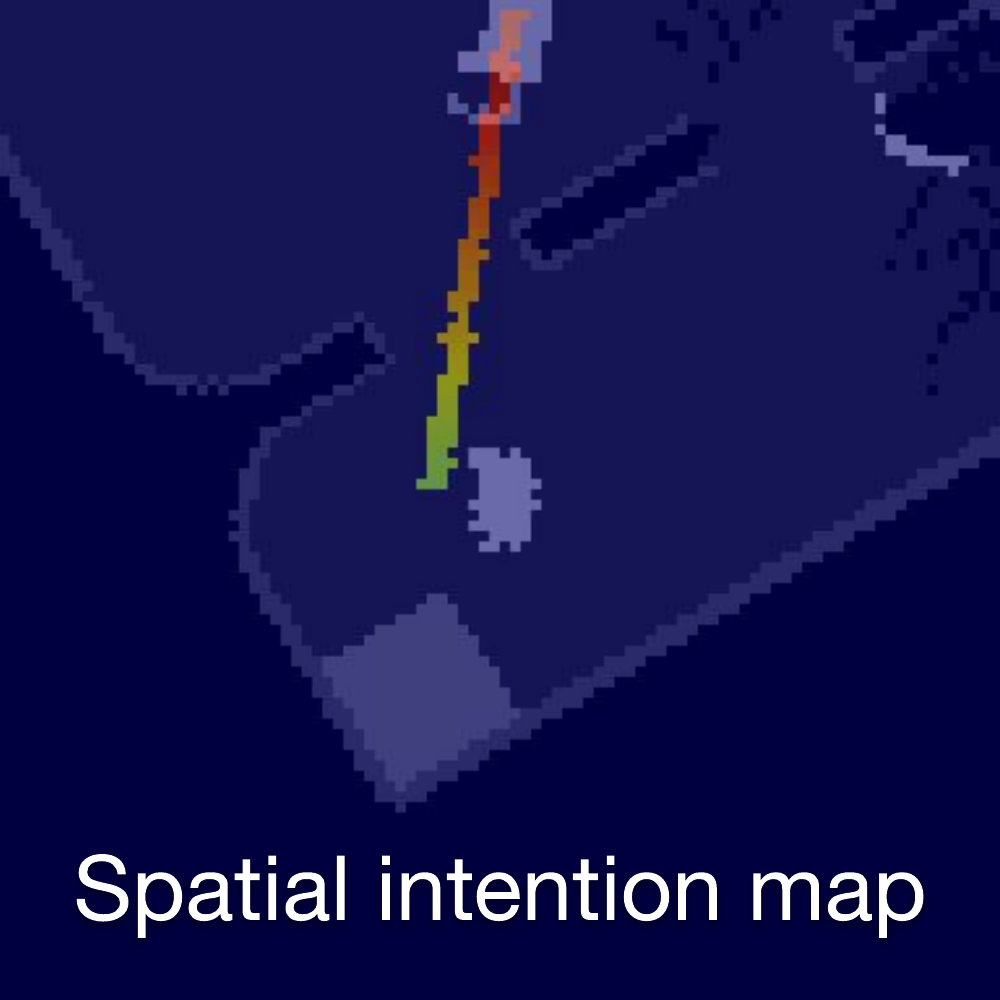} &
\includegraphics[width=0.248\columnwidth]{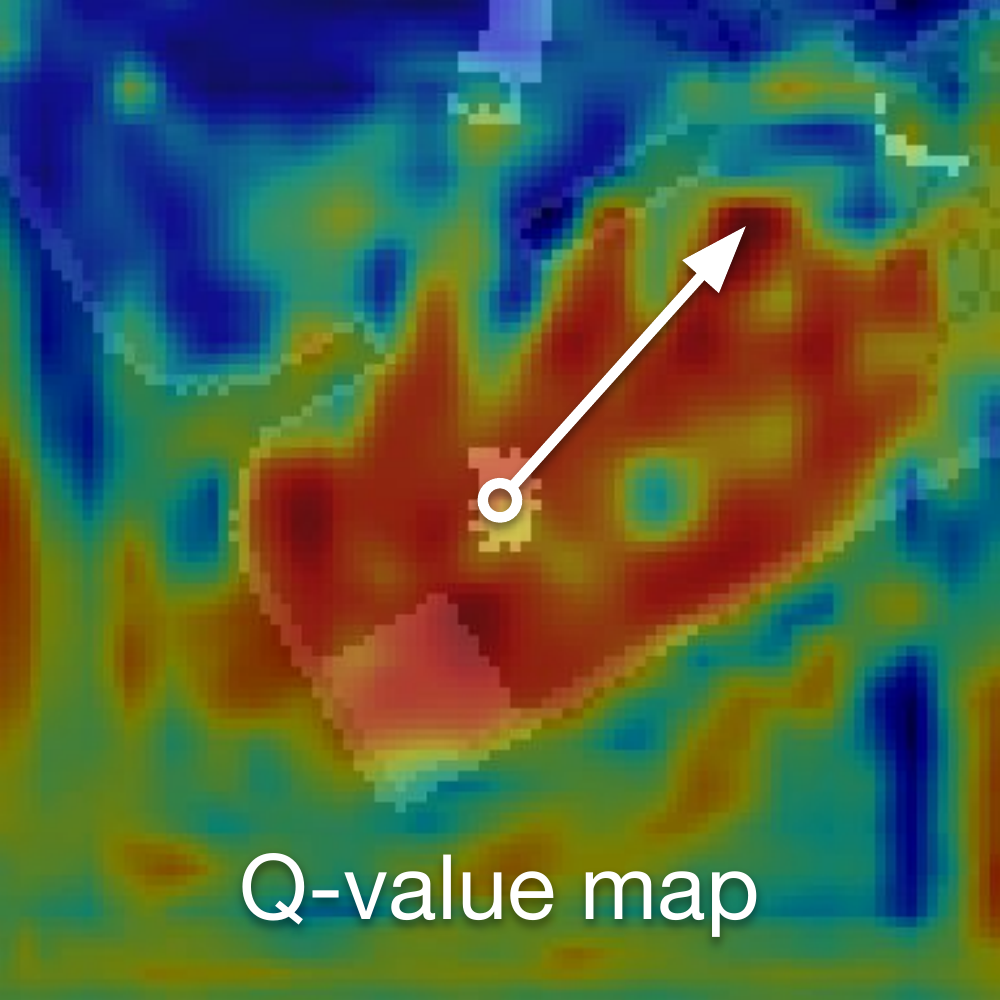} \\
\multicolumn{2}{c}{\small{Scenario 1}} & & \multicolumn{2}{c}{\small{Scenario 2}} \\
\end{tabular}
\end{center}
\vspace{-3mm}
\caption{\textbf{Coordinating to go through doorways.} In these scenarios, the current robot (center) is choosing a new action, while the other robot (top) is already moving. The other robot is headed towards the further (right) doorway in scenario 1 and the closer (left) doorway in scenario 2. In both scenarios, the Q-value map suggests the current robot should go through the unoccupied doorway.}
\label{fig:lifting-doorway-q-value-map}
\vspace{1mm}
\begin{center}
\setlength\tabcolsep{1pt}
\begin{tabular}{cc}
\includegraphics[width=0.492\columnwidth]{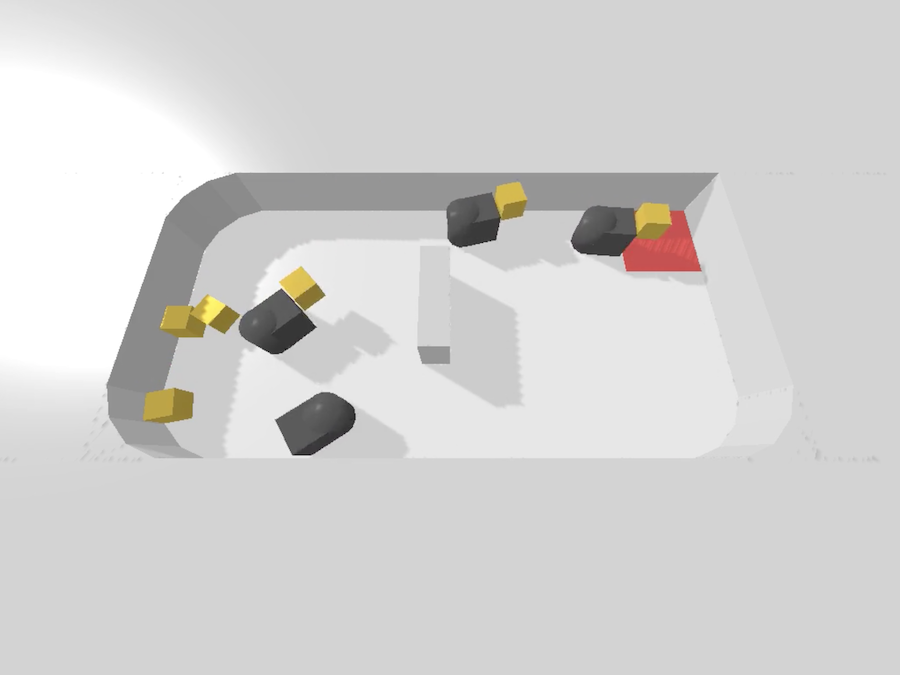} &
\includegraphics[width=0.492\columnwidth]{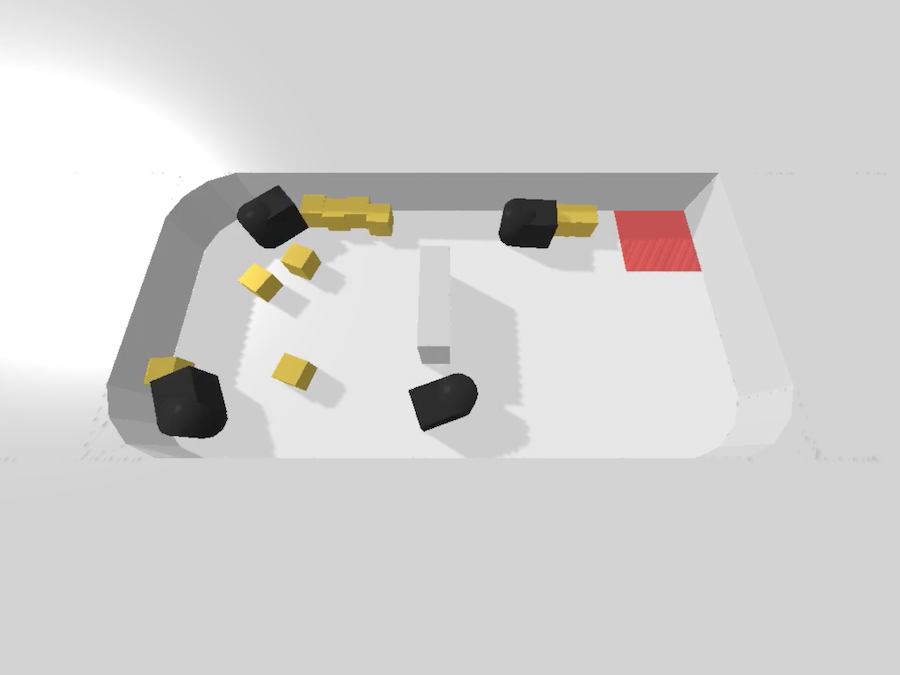} \\
\small{Lifting} & \small{Pushing} \\
\end{tabular}
\end{center}
\vspace{-3mm}
\caption{\textbf{Emergent foraging strategy.} With intention maps, both lifting and pushing teams learn an effective and efficient strategy in which they move single file in a clockwise circle around the center divider.}
\label{fig:emergent-small-divider}
\end{figure}

By inspecting output Q-value maps, we can validate that the spatial intention maps are being used in this way. 
Fig.~\ref{fig:lifting-doorway-q-value-map} examines a pair of scenarios requiring coordination between robots to pass through a shared doorway.
We find that the predicted Q-value maps for each robot assign higher Q-values to actions that are compatible with what the other robot intends to do.

We observe several interesting emergent strategies when training with intention maps.
For example, in the SmallDivider environment, both lifting and pushing robots consistently learn to move single file in a clockwise circle around the center divider (Fig.~\ref{fig:emergent-small-divider}).
In this emergent pattern, the robots move leftward through the bottom opening to find more objects, and rightward through the top opening to carry objects towards the receptacle.
By maintaining one-way traffic in each opening, the robots avoid having to pause and coordinate passing through an opening from opposite directions.
Without intention maps, we observe that this pattern still occasionally emerges, but is difficult to sustain.

\begin{table}
\caption{Search and Rescue Performance}
\vspace{-4mm}
\begin{center}
\begin{tabular}{l|cc}
\toprule
Environment & Ours & No intention maps \\
\midrule
SmallEmpty & \textbf{\x9.56 $\pm$ 0.28} & \x9.08 $\pm$ 0.45 \\
LargeEmpty & \textbf{19.52 $\pm$ 0.21} & 18.49 $\pm$ 0.72 \\
\bottomrule
\end{tabular}
\end{center}
\vspace{-6mm}
\label{tab:rescue}
\end{table}

\begin{figure}
\begin{center}
\setlength\tabcolsep{0pt}
\begin{tabular}{cc@{\hspace{1pt}}c@{\hspace{1pt}}cc}
\includegraphics[width=0.248\columnwidth]{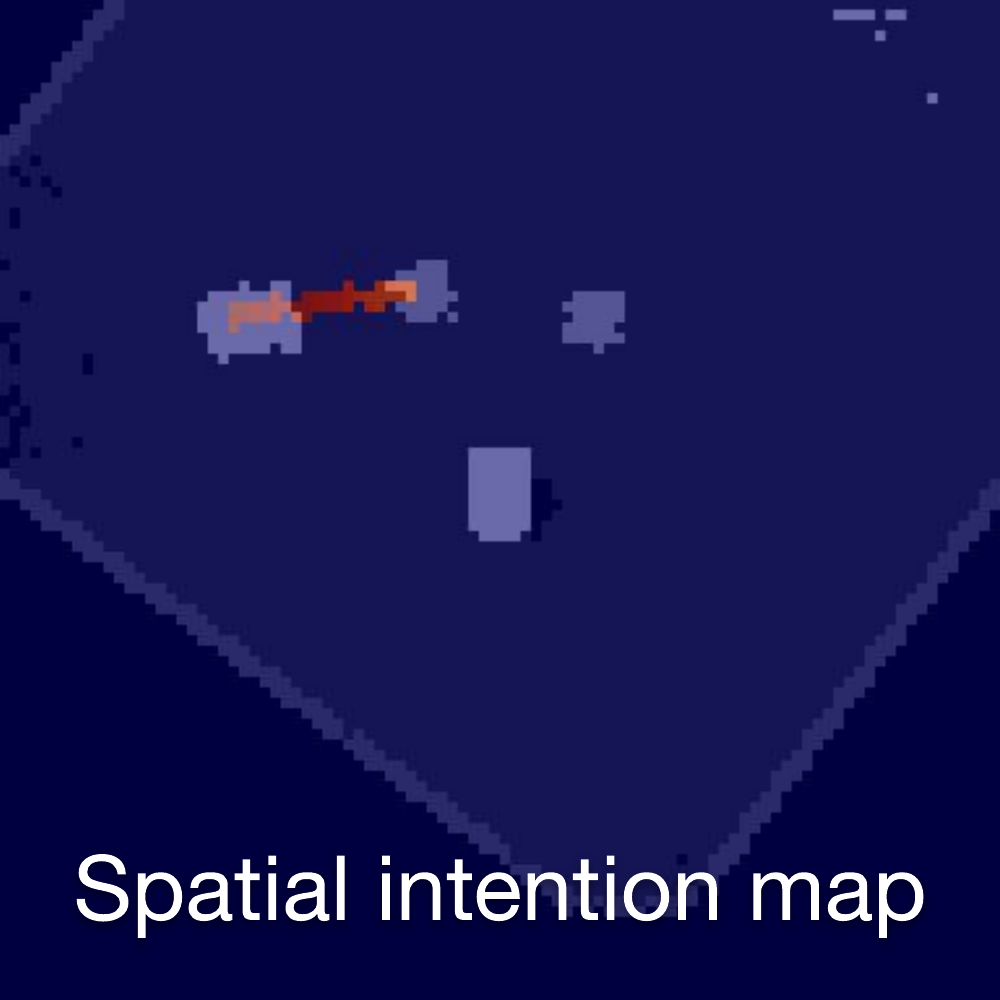} &
\includegraphics[width=0.248\columnwidth]{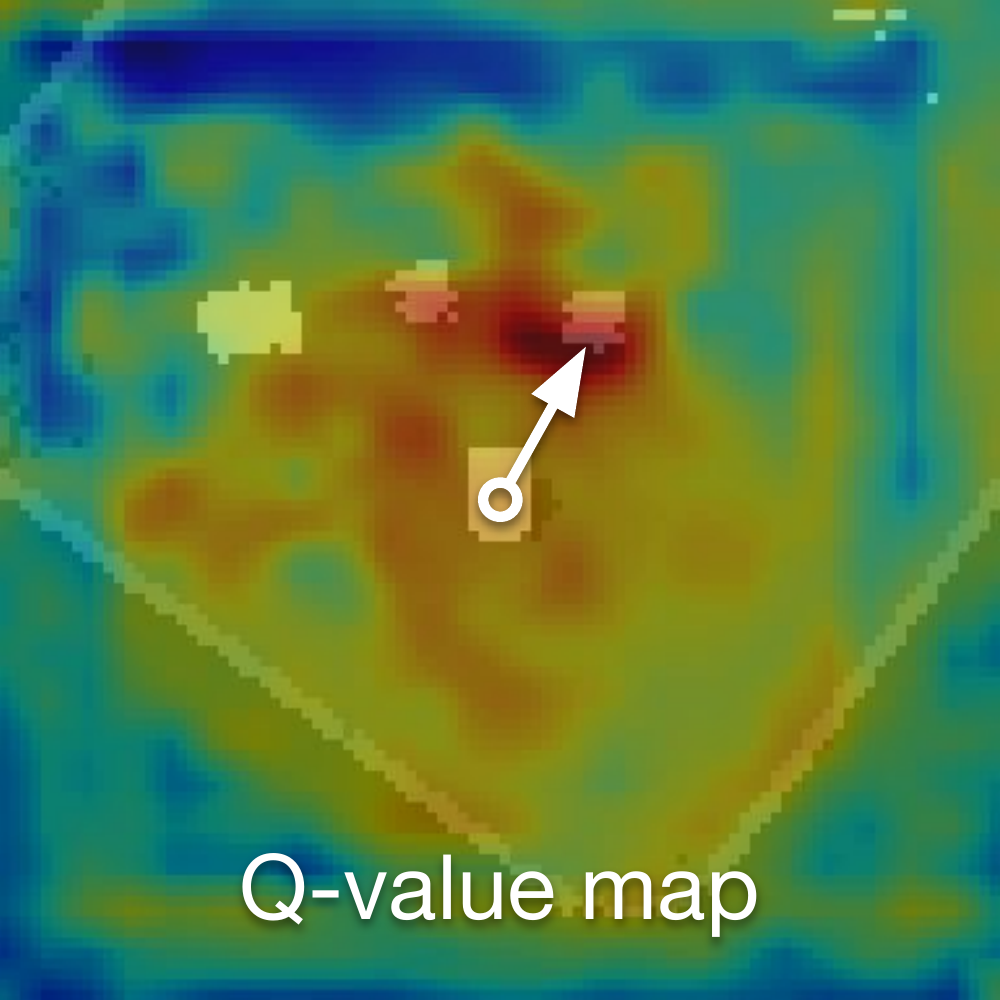} & &
\includegraphics[width=0.248\columnwidth]{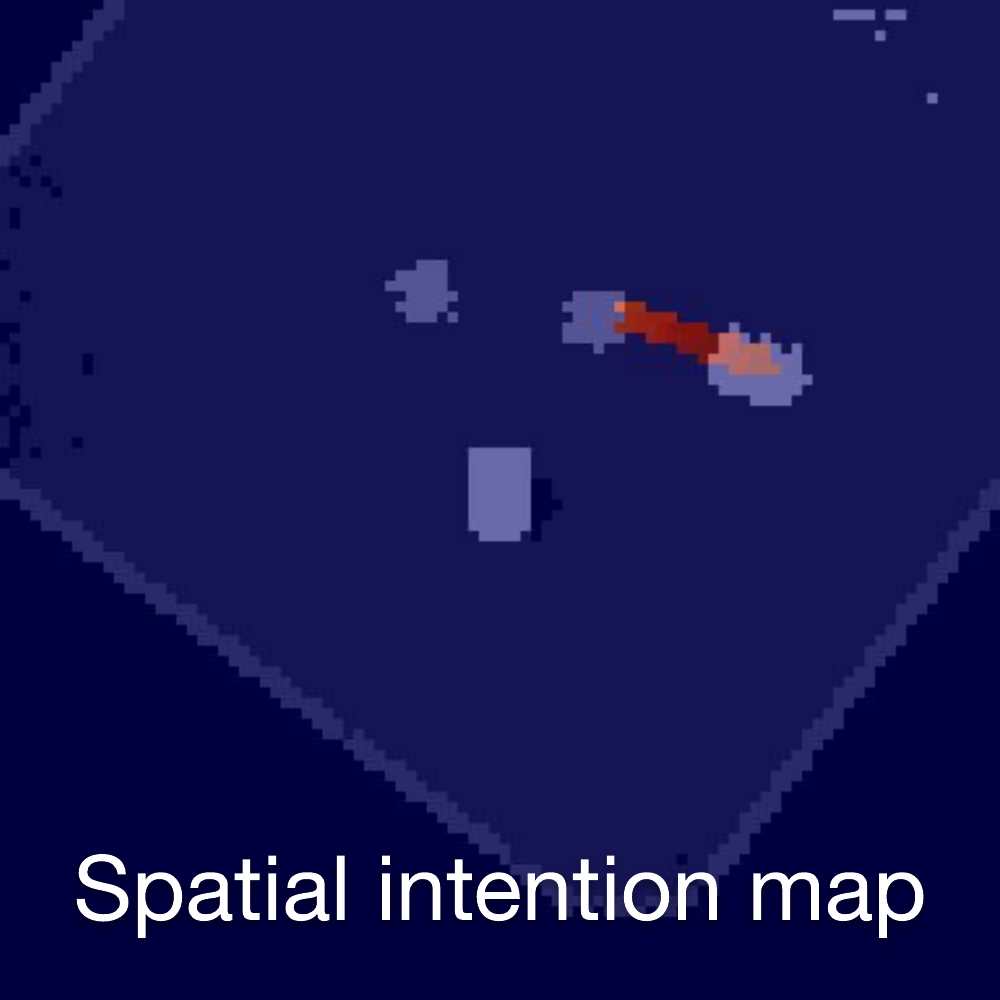} &
\includegraphics[width=0.248\columnwidth]{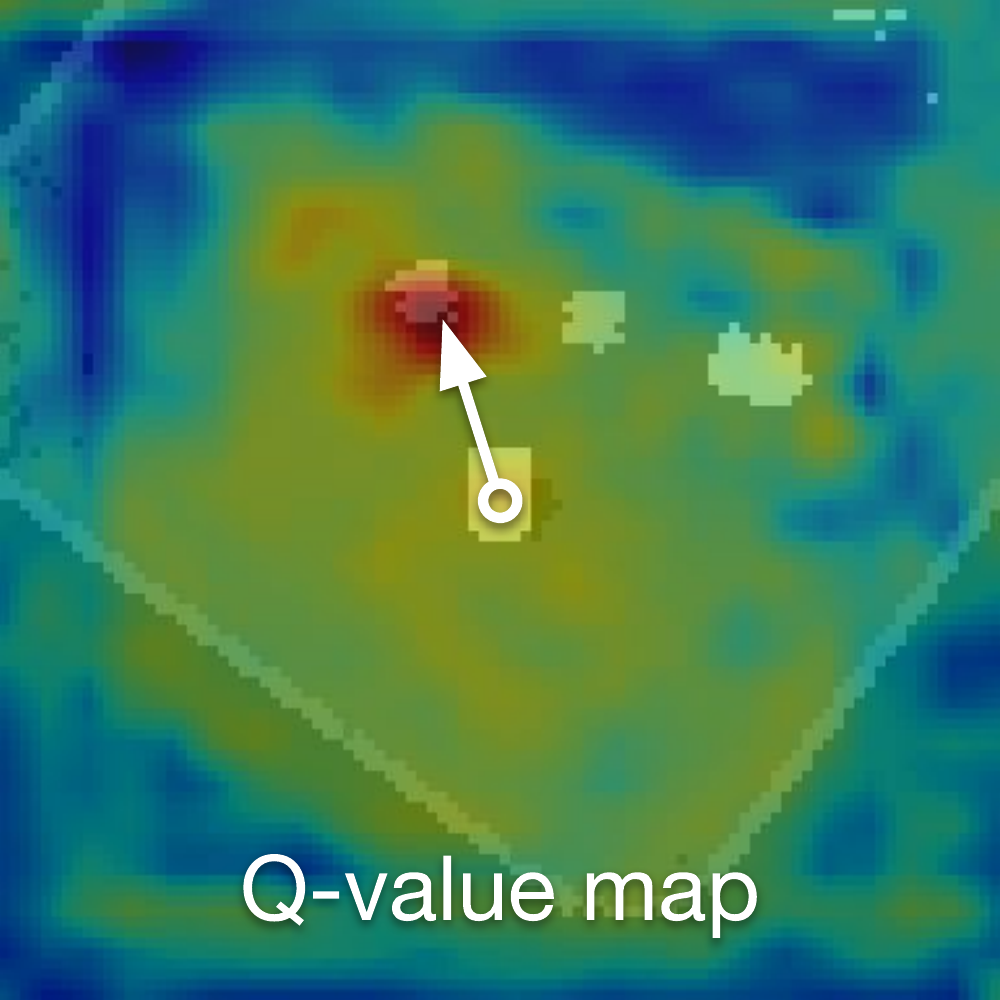} \\
\multicolumn{2}{c}{\small{Scenario 1}} & & \multicolumn{2}{c}{\small{Scenario 2}} \\
\end{tabular}
\end{center}
\vspace{-3mm}
\caption{\textbf{Coordinating to rescue objects.}
The current robot (center) is choosing a new action, while the other robot is already moving. They are trying to rescue the two objects (small squares on top).
The other robot intends to rescue the left object in scenario 1, and the right object in scenario 2.
In both scenarios, the Q-value map suggests the current robot should rescue the opposite object, to avoid overlapping of efforts.}
\label{fig:rescue-q-value-map}
\vspace{1mm}
\begin{center}
\setlength\tabcolsep{1pt}
\begin{tabular}{cc}
\includegraphics[width=0.492\columnwidth]{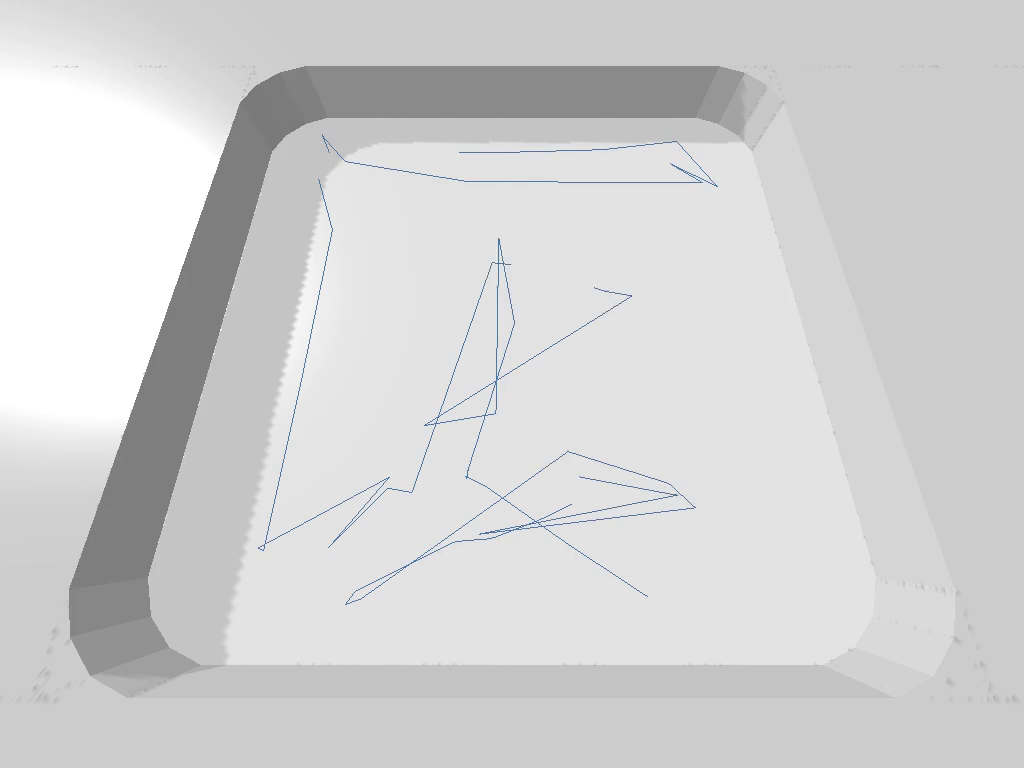} &
\includegraphics[width=0.492\columnwidth]{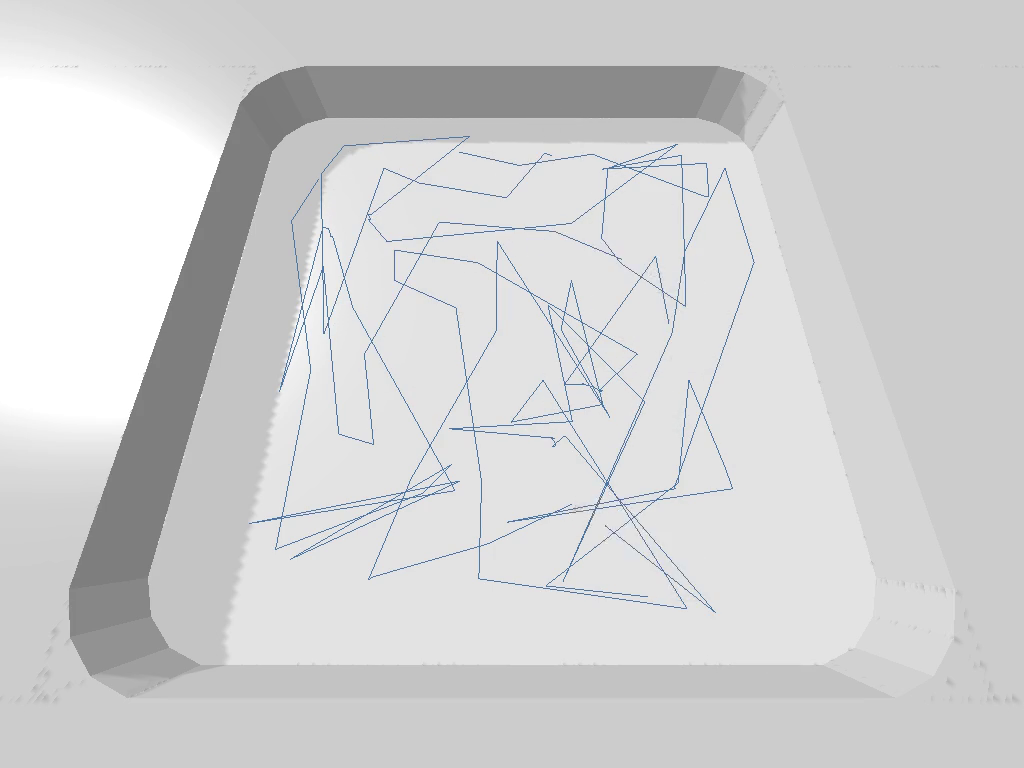} \\
\small{Ours} & \small{No intention maps} \\
\end{tabular}
\end{center}
\vspace{-3mm}
\caption{\textbf{Search and rescue team efficiency.} Movement trajectories (blue) over an episode show that rescue robots finish their task more efficiently when intention maps are used. Without intention maps, the robots are unable to coordinate as well since they do not know the intentions of other robots.}
\label{fig:rescue-trajectories}
\vspace{1mm}
\begin{center}
\setlength\tabcolsep{1pt}
\begin{tabular}{cc}
\includegraphics[width=0.492\columnwidth]{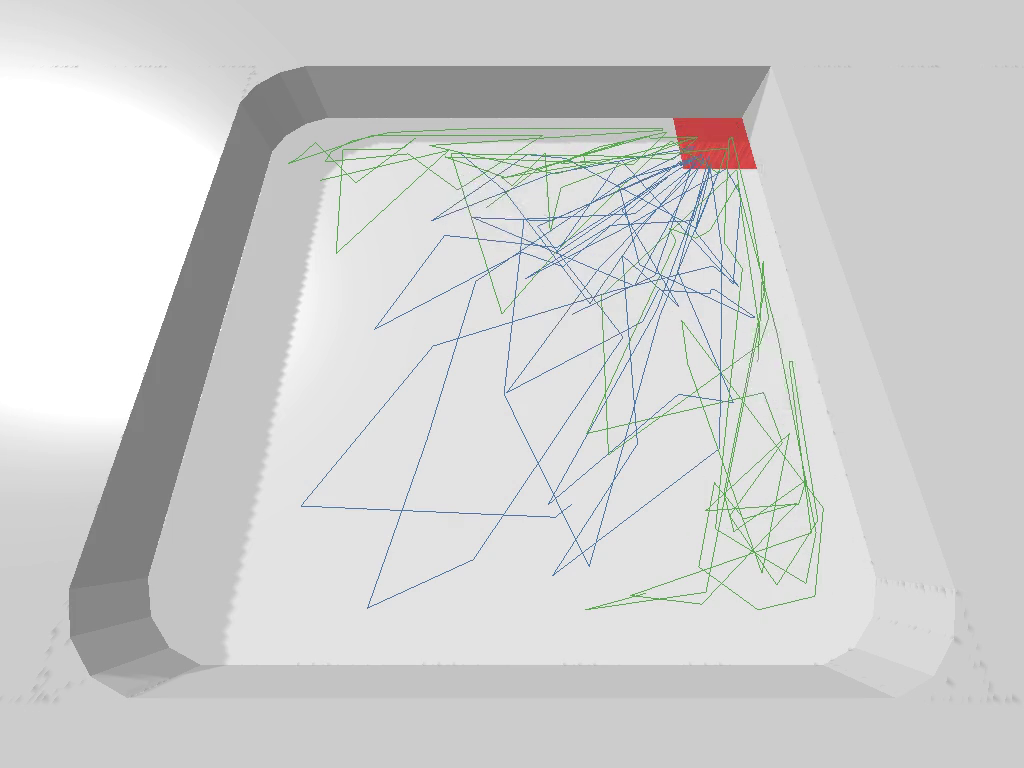} &
\includegraphics[width=0.492\columnwidth]{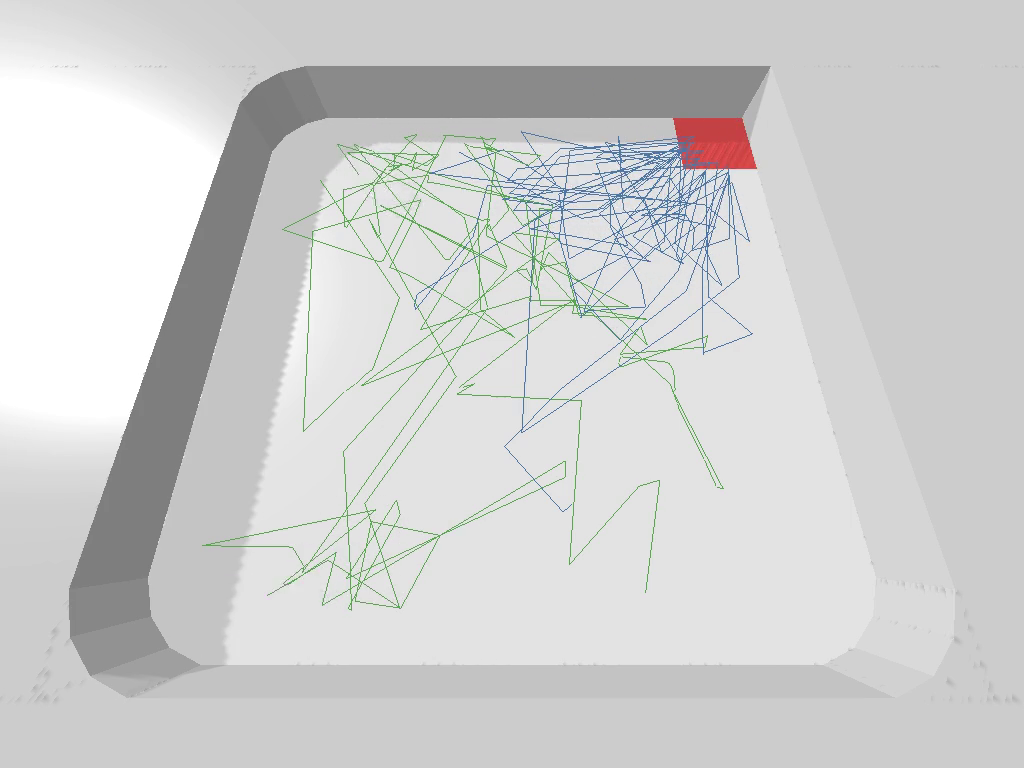} \\
\small{Lifting + Pushing} & \small{Lifting + Throwing} \\
\end{tabular}
\end{center}
\vspace{-3mm}
\caption{\textbf{Emergent division of labor for heterogenous teams.} When we train heterogeneous teams with spatial intention maps, we see from movement trajectories that a natural division of labor emerges (lifting trajectories are blue, pushing/throwing trajectories are green).
Notice that there is almost no overlap between blue and green trajectories in either image.
We see that pushing robots focus on objects along the wall since those can be pushed much more reliably, while throwing robots focus on faraway objects since they can throw them backwards a long distance towards the receptacle.}
\vspace{-6mm}
\label{fig:heterogeneous-trajectories}
\end{figure}

\mysubsection{Search and rescue task.}
We then investigate whether spatial intention maps are also helpful for the search and rescue task. Tab.~\ref{tab:rescue} shows results for teams of 4 rescue robots in the SmallEmpty and LargeEmpty environments.
Both quantitative and qualitative results indicate that robots without intention maps are less efficient. They tend to move back and forth unproductively as multiple robots try to rescue the same object, especially towards the end of an episode (when there are only a few objects left).
Since the search and rescue task by its nature finishes much more quickly than the foraging task, we are able to see this behavior very clearly in the movement trajectories (Fig.~\ref{fig:rescue-trajectories}).
By contrast, robots that use intention maps know the intentions of other robots, and can choose actions that avoid overlapping with others, leading to better distribution of robots throughout the environment.
This emerging behavior can be seen clearly in visualizations of Q-value maps of trained policies.
For example, in Fig.~\ref{fig:rescue-q-value-map}, we see that the locations corresponding to objects another robot already intends to rescue are generally assigned lower Q-values.
These results indicate that spatial intention maps can lead to more efficient execution of the rescue task by helping agents avoid overlapping in their efforts.

\begin{table*}
\caption{Comparisons and Ablations}
\vspace{-6mm}
\begin{center}
\setlength\tabcolsep{4pt}
\resizebox{\textwidth}{!}{%
\begin{tabular}{l|cccc|cc|cc|cc}
\toprule
& \multicolumn{6}{c|}{Explicit communication} & \multicolumn{4}{c}{Implicit communication} \\
\midrule
& \multicolumn{4}{c|}{Intention maps} & \multicolumn{2}{c|}{Intention channels} & \multicolumn{2}{c|}{Baselines} & \multicolumn{2}{c}{Predicted intention} \\
Environment & Ours & Binary & Line & Circle & Spatial & Nonspatial & No intention & History maps & No history & With history \\
\midrule
SmallEmpty & \x9.54 $\pm$ 0.16 & \x9.25 $\pm$ 0.27 & \x9.56 $\pm$ 0.15 & \x9.19 $\pm$ 0.33 & \x9.33 $\pm$ 0.43 & \x8.38 $\pm$ 0.52 & \x7.92 $\pm$ 0.86 & \x9.29 $\pm$ 0.16 & \x8.95 $\pm$ 0.32 & \x9.05 $\pm$ 0.30 \\
SmallDivider & \x9.44 $\pm$ 0.46 & \x9.28 $\pm$ 0.49 & \x8.98 $\pm$ 0.89 & \x9.55 $\pm$ 0.16 & \x9.47 $\pm$ 0.37 & \x8.73 $\pm$ 0.85 & \x8.07 $\pm$ 0.75 & \x9.20 $\pm$ 0.61 & \x8.69 $\pm$ 0.90 & \x9.11 $\pm$ 0.43 \\
LargeEmpty & 18.86 $\pm$ 0.85 & 19.51 $\pm$ 0.47 & 19.43 $\pm$ 0.17 & 17.41 $\pm$ 3.75 & 18.36 $\pm$ 0.94 & 18.15 $\pm$ 0.54 & 15.58 $\pm$ 3.80 & 17.88 $\pm$ 1.56 & 18.18 $\pm$ 1.32 & 18.29 $\pm$ 1.45 \\
LargeDoors & 19.57 $\pm$ 0.25 & 18.38 $\pm$ 1.98 & 17.84 $\pm$ 1.16 & 17.89 $\pm$ 1.43 & 18.43 $\pm$ 0.52 & 14.07 $\pm$ 1.89 & 13.96 $\pm$ 2.32 & 16.14 $\pm$ 2.15 & 17.84 $\pm$ 1.55 & 18.81 $\pm$ 0.94 \\
LargeTunnels & 19.00 $\pm$ 0.47 & 18.95 $\pm$ 0.75 & 18.11 $\pm$ 1.96 & 19.51 $\pm$ 0.42 & 18.65 $\pm$ 0.87 & 12.43 $\pm$ 1.73 & 11.89 $\pm$ 5.96 & 18.08 $\pm$ 1.35 & 18.74 $\pm$ 0.81 & 18.07 $\pm$ 1.89 \\
LargeRooms & 19.52 $\pm$ 0.38 & 18.59 $\pm$ 0.99 & 18.84 $\pm$ 0.96 & 19.51 $\pm$ 0.31 & 19.15 $\pm$ 0.57 & 17.55 $\pm$ 0.30 & 16.56 $\pm$ 1.53 & 17.84 $\pm$ 0.58 & 18.97 $\pm$ 0.34 & 19.35 $\pm$ 0.19 \\
\bottomrule
\end{tabular}}
\end{center}
\vspace{-6mm}
\label{tab:communication}
\end{table*}

\mysubsection{Heterogeneous teams.} 
We also investigate whether spatial intention maps are useful for heterogeneous teams of robots.
The bottom five rows of Tab.~\ref{tab:foraging} show results on the foraging task for teams of 2 lifting and 2 pushing robots (2L+2P), and for teams of 2 lifting and 2 throwing robots (2L+2T). While the lifting robot is generally versatile, the pushing robot is particularly adept at pushing objects along walls, while the throwing robot can throw objects long distances backwards. The unique capabilities of the different robot types are complementary to each other, and indeed, we find that when we train these heterogeneous teams, a natural division of labor emerges, as shown in Fig.~\ref{fig:heterogeneous-trajectories}. This specialization generally happens with or without spatial intention maps, but we find that when training without intention maps, some of the robots in the team may learn to perpetually wander around without doing anything useful, or the team may never learn to fully complete the task. They likely develop this conservative behavior since they are unable to coordinate and avoid collisions without knowing each other's intentions. By contrast, when training with intention maps, heterogeneous teams are more efficient and productive.

\subsection{Comparisons and Ablations}
\label{sec:experiments-comparisons-ablations}

In this section, we perform comparisons and ablations across a wide range of communication variants. For all variants, we run experiments on the foraging task using homogeneous teams of 4 lifting robots in all six environments.

\mysubsection{Comparison to nonspatial intentions.} We compare spatial intention maps to a nonspatial encoding of intentions, in which we add two channels per robot (sorted closest robot to furthest) and tile them with the $x$ and $y$ values of each robot's intended target location.
This nonspatial encoding conveys to the network the $(x , y)$ target of other robots' intended destination in the best way we can imagine without a spatially varying map.
As shown in Tab.~\ref{tab:communication}, we find that providing the intentions encoded in a nonspatial format results in much worse performance, suggesting that it is important to use a spatial encoding for agent intentions.

\mysubsection{Encoding of spatial intentions.} Our method encodes intended paths (with linearly ramped values) in a 2D map. Here we investigate four alternative encodings:
(i) a binary encoding of the intention map where the linear ramp of values is replaced with a binaryon-path off-path map,
(ii) a line encoding that replaces the path encoding with a simple straight line between each agent to its intended target location (which may not necessarily reflect the agent's true trajectory),
(iii) a circle encoding, where a circle marks each agent's target location (this variant does not associate each agent with its target location), and
(iv) an encoding that decomposes the circle intention map into multiple channels (``spatial intention channels'' in Tab.~\ref{tab:communication}), one per robot, sorted in order from closest robot to furthest (thus allowing a way to associate agents with intentions).
Results in Tab.~\ref{tab:communication} show that the binary and line variants generally perform on par with our method, while both circle variants are worse overall (but still better than without intention maps).
These results suggest that providing any spatial encoding of intention is highly beneficial to multi-agent coordination, as long as there is a clear visual association between agents and their intentions.

\mysubsection{Predicting spatial intention maps.} We further explore whether agents can learn to predict useful intention maps, either with heuristics or deep networks.  We explore three possible methods:
(i) predicted intention, which involves training an additional fully convolutional network to predict intention maps from the state representation so that they can be used during execution with no communication,
(ii) history maps, where instead of encoding intention, the recent trajectory history of other agents are encoded into a map (assumes robots can track each other's poses without communicating), and
(iii) a combination of (i) and (ii), where an additional network is trained to predict the intention map from a state representation augmented with the history map.
Results in Tab.~\ref{tab:communication} show that in isolation, (i) and (ii) do not improve performance as much as spatial intention maps do. However, we find that the last variant (iii) which combines history maps with predicted intention achieves performance (Tab.~\ref{tab:communication}) almost on par with explicit communication using spatial intention maps.  This result is significant in that it provides a method for robot teams to leverage the benefits of spatial intention maps to coordinate without explicit communication, albeit at the cost of extra computation.

\subsection{Real Robot Experiments}
\label{sec:experiments-real}

As a final test, we take the best policies from our simulation experiments and run them directly on the real robots by building upon the physical setup used in \cite{wu2020spatial}.
We use fiducial markers to estimate the poses of robots and objects, which allows us to mirror the real setup inside the simulation.
This means we can execute the trained policies (without fine-tuning) on the real robots as if they were moving around in the simulation.
During tests with 4 lifting robots in a mirror of the SmallEmpty environment, the robots are able to gather all 10 objects within 1 minute and 56 seconds (on average over 5 episodes).
For video results across a wide range of environments, please see the supplementary material at \url{https://spatial-intention-maps.cs.princeton.edu}.

\section{Conclusion}

In this work, we propose \emph{spatial intention maps} as a new way of communicating intention for improved collaboration in multi-agent RL.
In our framework, intentions are spatially encoded in a 2D map, allowing vision-based RL agents to spatially reason about intentions in the same domain as the states and actions.
During experiments, we observe that spatial intention maps provide significant gains in performance across a broad range of multi-agent environments, and help robot teams learn emergent collaborative behaviors such as avoiding collisions, coordinating passage through bottlenecks, and distributing throughout an environment.

\section*{Acknowledgments}
We thank Naomi Leonard, Anirudha Majumdar, Naveen Verma, and Anish Singhani for fruitful technical discussions. This work was supported by Princeton School of Engineering, as well as NSF under IIS-1815070 and DGE-1656466.

\newpage
\bibliographystyle{IEEEtran}
\bibliography{IEEEabrv,references}

\clearpage
\section*{Appendix}

\subsection{Additional Reinforcement Learning Details}

We model our task as a Markov decision process (MDP) from the perspective of each individual agent. Specifically, given state $s_t$ at time $t$, the agent takes an action $a_t$ following policy $\pi(s_t)$, and arrives at a new state $s_{t+1}$ while receiving reward $r_t$.
In Q-learning, the goal is to find some optimal policy $\pi^*$ that selects actions maximizing the Q-function $Q(s_t,a_t)=\sum_{i=t}^{\infty}\gamma^{i-t} r_i$, which represents the discounted sum of future rewards.
We train our policies using deep Q-learning (DQN)~\cite{mnih2015human}, which approximates the Q-function using a neural network and uses a policy that greedily selects actions maximizing the Q-function: $\pi(s_t) = \argmax_{a_t} Q_\theta(s_t,a_t)$, where $\theta$ refers to the parameters of the neural network.
We train using the double DQN learning objective~\cite{van2016deep} with smooth L1 loss. Formally, at each training iteration $i$, we minimize
\resizebox{\columnwidth}{!}{$\mathcal{L}_i = |r_t + \gamma Q_{\theta_i^-}(s_{t+1},\argmax_{a_{t+1}}{Q_{\theta_i}(s_{t+1},a_{t+1})})-Q_{\theta_i}(s_t,a_t)|$}
where $(s_t,a_t,r_t,s_{t+1})$ is a transition uniformly sampled from the replay buffer, and $\theta^-$ refers to the parameters of the DQN target network.

\subsection{Additional Training Details}

We train DQN using SGD for either 160k or 240k timesteps (depending on the robot types in the team), with batch size 32, learning rate 0.01, momentum 0.9, and weight decay 0.0001. Norms of gradients are clipped to 100 during training.
We use a replay buffer of size 10k, constant discount factor $\gamma=0.85$, train frequency of 4 (train the policy network every 4 timesteps), and we update the target network every 1,000 timesteps.
Episodes end after all objects have been removed from the environment, or after 400 consecutive steps in which no object has been removed.
Training takes approximately 6 hours on a single Nvidia Titan Xp GPU (for the lifting team).

\mysubsection{Network architecture.}
We model our Q-function $Q_\theta$ with a ResNet-18~\cite{he2016deep} backbone, which we transform into a fully convolutional network by removing the AvgPool and fully connected layers and adding three 1x1 convolution layers interleaved with bilinear upsampling layers. Unlike in \cite{wu2020spatial}, we apply BatchNorm after convolutional layers.

\mysubsection{Exploration.}
We run a random policy for a small number of timesteps (1/40 of the total number of timesteps) in order to fill the replay buffer with some initial experience before we start training the policy network.
We use $\epsilon$-greedy exploration, in which we linearly anneal the exploration fraction from 1 to 0.01 during the first 1/10 of training.

\mysubsection{Multi-agent.}
We use one buffer/policy per robot type during training. This means that homogeneous teams pool all transitions into one replay buffer and train one shared policy, while heterogeneous teams share a buffer/policy within each group of robots (of one robot type). For example, the lifting and pushing team trains with two buffers/policies.
For decentralized execution, trained policies are independently and asynchronously run on each individual agent.

\mysubsection{Search and rescue.}
We made a few small modifications to the foraging task for search and rescue.
Since the receptacle is not relevant, we remove it for this task, and also remove the corresponding channel (shortest path distances to the receptacle) from the state representation.
To reflect the shorter term nature of this simpler task, we lower the discount factor $\gamma$ to 0.35 and reduce the number of training timesteps to 15k.

\mysubsection{Predicted intention.}
To predict intention maps, we train an additional fully convolutional network in tandem with our policy.
We use the same network architecture as our policy network (but with a sigmoid layer at the end) and train with binary cross entropy loss (supervised on our intention map).
The policy network uses our intention map as input for the first 9/10 of training and switches to the predicted intention map for the last 1/10 of training (as well as execution).

\begin{figure}
\begin{center}
\setlength\tabcolsep{0pt}
\begin{tabular}{cc@{\hspace{1pt}}c@{\hspace{1pt}}cc}
\includegraphics[width=0.248\columnwidth]{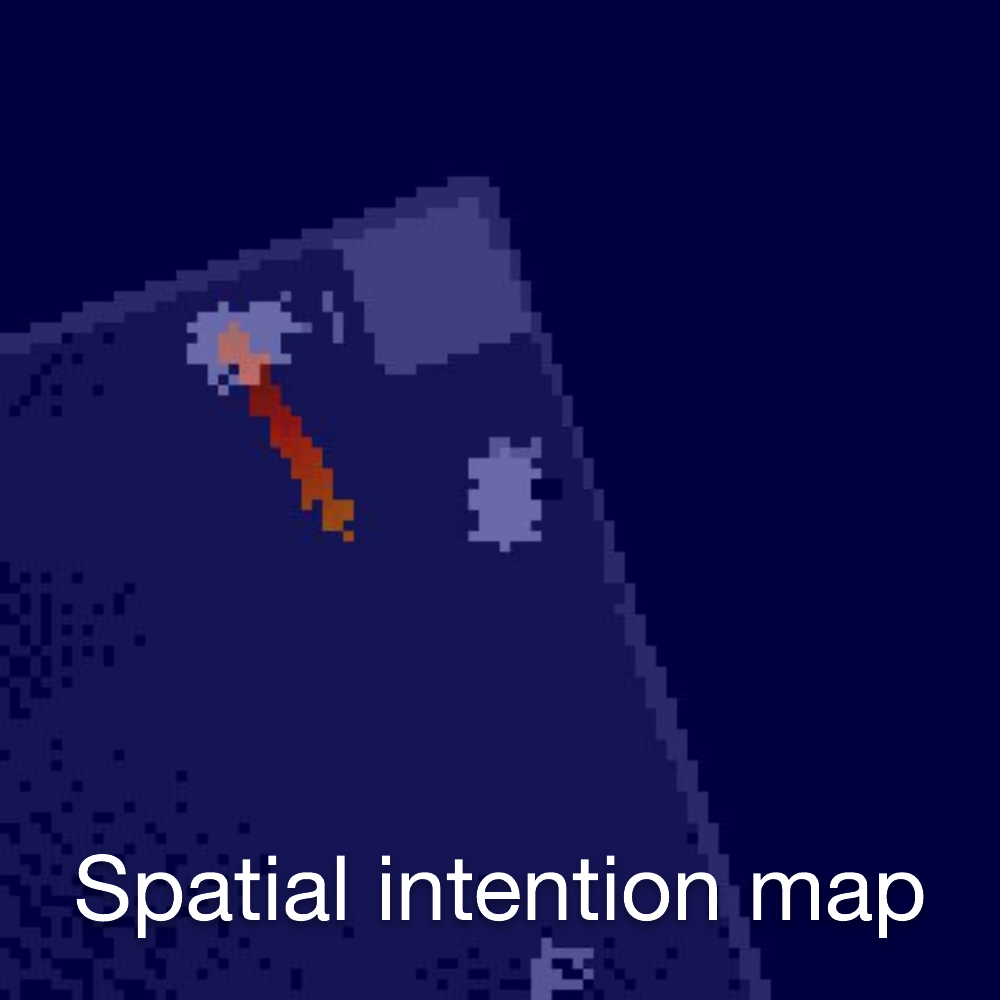} &
\includegraphics[width=0.248\columnwidth]{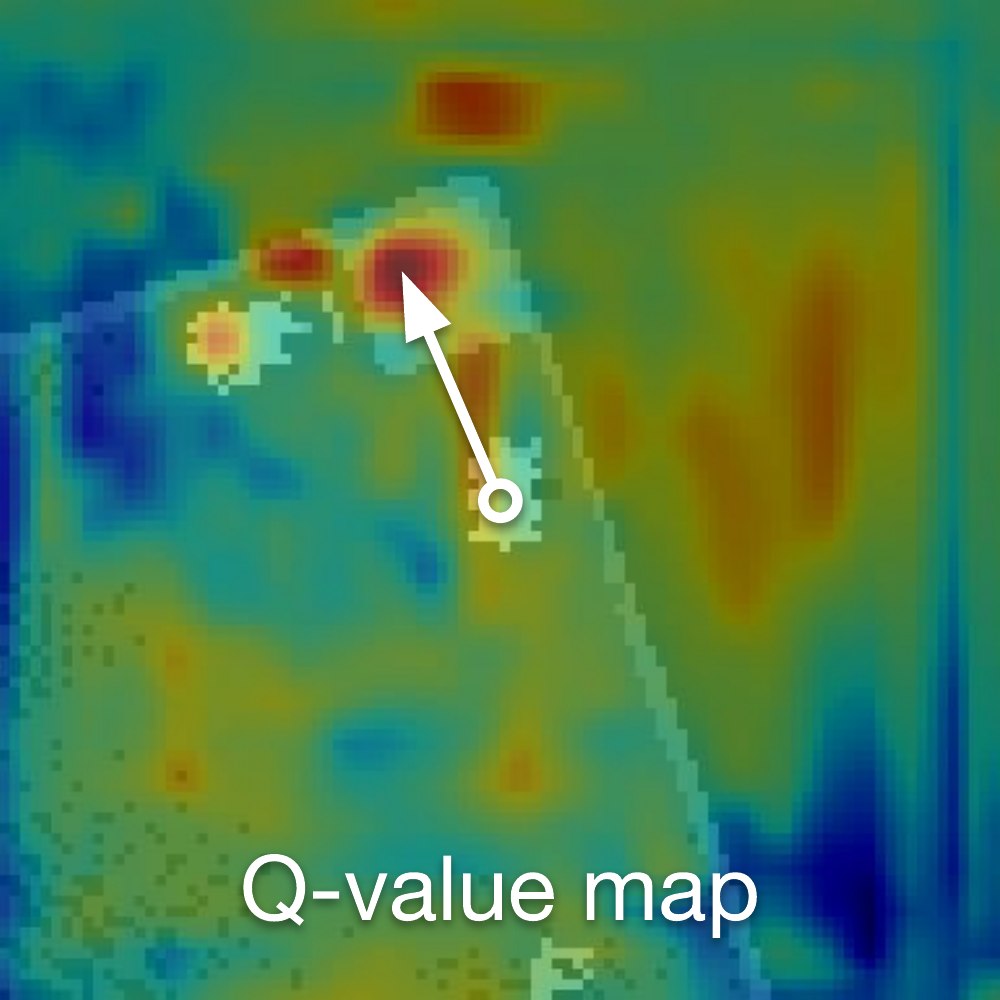} & &
\includegraphics[width=0.248\columnwidth]{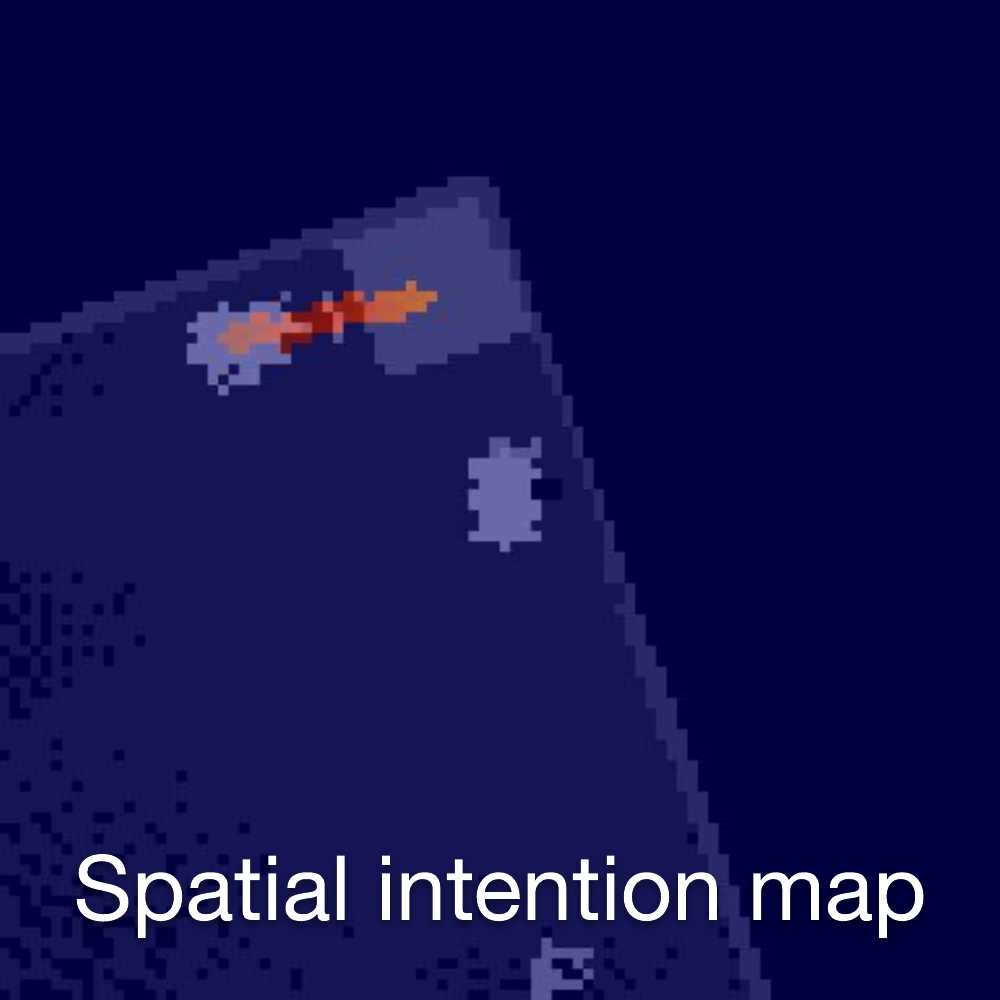} &
\includegraphics[width=0.248\columnwidth]{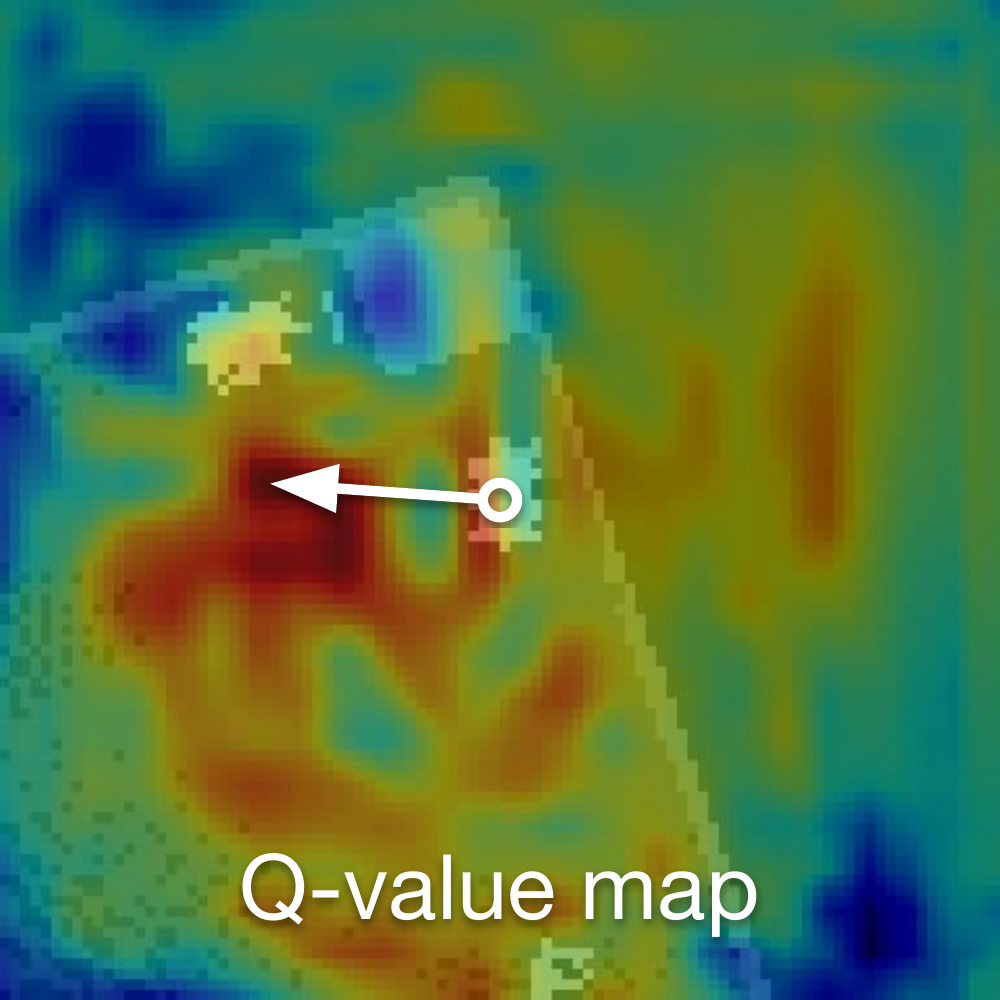} \\
\multicolumn{2}{c}{\small{Scenario 1}} & & \multicolumn{2}{c}{\small{Scenario 2}} \\
\end{tabular}
\end{center}
\vspace{-2mm}
\caption{\textbf{Coordinating to go towards the receptacle.}
In these scenarios, the current robot (center) is choosing a new action, while the other robot (left) is already moving.
The other robot is moving away from the receptacle in scenario 1, and towards the receptacle in scenario 2.
The Q-value map suggests that the current robot should drop off the object in the receptacle only if it does not conflict with the other robot. Otherwise, it suggests that the current robot should move left (and wait for the other robot to finish).
}
\label{fig:lifting-receptacle-q-value-map}
\end{figure}

\begin{figure}
\begin{center}
\setlength\tabcolsep{1pt}
\begin{tabular}{ccc}
\includegraphics[width=0.325\columnwidth]{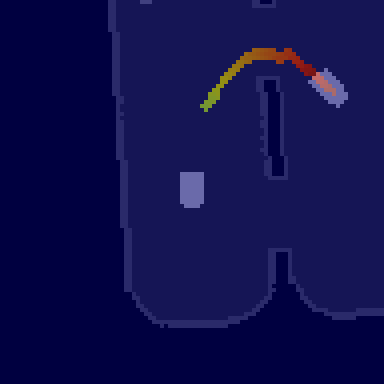} &
\includegraphics[width=0.325\columnwidth]{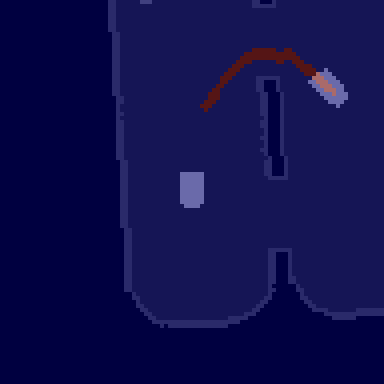} &
\includegraphics[width=0.325\columnwidth]{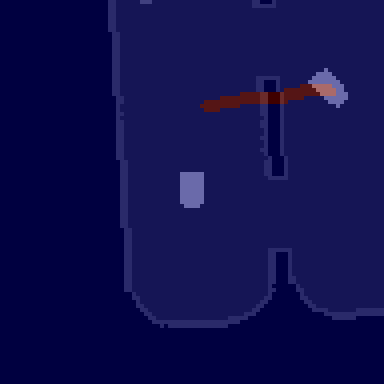} \\
\small{(a) Ours} & \small{(b) Binary} & \small{(c) Line} \\
\vspace{-2mm} \\
\includegraphics[width=0.325\columnwidth]{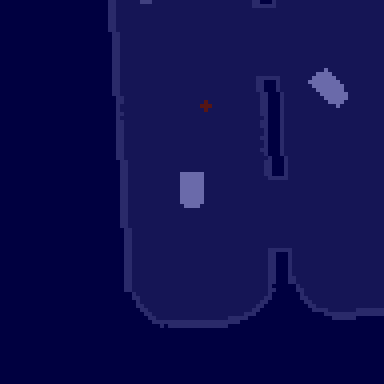} &
\includegraphics[width=0.325\columnwidth]{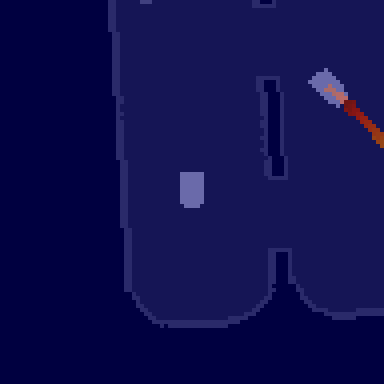} &
\includegraphics[width=0.325\columnwidth]{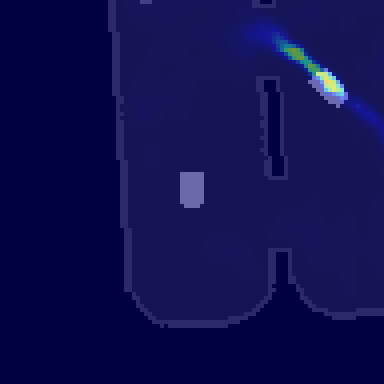} \\
\small{(d) Circle} & \small{(e) History} & \small{(f) Predicted} \\
\end{tabular}
\end{center}
\vspace{-2mm}
\caption{\textbf{Communication variants.} Here we illustrate the different communication variants we experimented with. In each image, the current robot (center) is choosing a new action, while the other robot (right) is moving towards the left. The images show (a) our method, (b-d) variants of spatial intention maps, (e) history maps, and (f) predicted intention maps.}
\vspace{-4mm}
\label{fig:variants}
\end{figure}

\subsection{Additional Figures}

Fig.~\ref{fig:lifting-receptacle-q-value-map}
shows output Q-value maps for another pair of scenarios (lifting robots coordinate to go towards the receptacle) to further validate that spatial intention maps are used by trained policies in a meaningful way.
Fig.~\ref{fig:variants} shows the different communication variants we study in Sec.~\ref{sec:experiments-comparisons-ablations}.

\end{document}